\documentclass[journal]{IEEEtran}
\ifCLASSINFOpdf
\else
\fi

\usepackage{multirow} 
\usepackage{array}

\usepackage{amsmath}
\usepackage{amssymb}
\usepackage{mathtools}
\usepackage{amsthm}
\usepackage{enumitem}
\usepackage{xcolor}

\usepackage{subfiles} %
\usepackage{makecell} %
\usepackage{multirow} %
\usepackage{subcaption}
\usepackage{color,xcolor}
\usepackage{graphicx}
\usepackage{svg}
\usepackage{booktabs}
\usepackage{makecell}
\usepackage{colortbl}
\usepackage{pifont}%
\usepackage{algorithm} %
\usepackage{listings}
\usepackage{algorithmic} %
\usepackage[switch]{lineno}
\usepackage{amssymb} %

\usepackage[misc]{ifsym}

\usepackage[normalem]{ulem}

\usepackage{tcolorbox}
\usepackage{hyperref}
\usepackage{longtable}

\usepackage{amssymb}
\usepackage{amsmath}
\usepackage{booktabs} 
\usepackage[table]{xcolor}
\usepackage{multirow}

\theoremstyle{plain}

\theoremstyle{definition}

\theoremstyle{remark}

\usepackage[textsize=tiny]{todonotes}
\usepackage{xspace}
\definecolor{shapecolor}{rgb}{0.0,0.5,0.0}

\definecolor{arylideyellow}{rgb}{0.91, 0.84, 0.42}

\def\etal{\emph{et al.}}

\newcommand{\sysName}{\texttt{MobiDiary}\xspace}

\usepackage{orcidlink}

\begin{document}
%
\title{MobiDiary: Autoregressive Action Captioning with Wearable Devices and Wireless Signals}
%
%
%

\author{Fei Deng\orcidlink{0009-0005-3319-7561},~
        Yinghui He\orcidlink{0000-0002-5560-1327},~
        Chuntong Chu\orcidlink{0009-0004-1904-8538},~
        Ge Wang\orcidlink{0009-0005-9104-5056},~
        Han~Ding\orcidlink{0000-0002-5274-7988},~\IEEEmembership{Senior~Member,~IEEE},~\\
        Jinsong Han\orcidlink{0000-0001-5064-1955},~\IEEEmembership{Senior~Member,~IEEE},~
        Fei Wang\orcidlink{0000-0002-0750-6990}
\thanks{Under Review. This work was supported by the National Natural Science Foundation of China under grants  62302383, 62372400, 62372365, 62472346.}
\thanks{Fei Deng~(email: dengfei@stu.xjtu.edu.cn), Chuntong Chu~(email: chuntong@stu.xjtu.edu.cn), and Fei Wang~(email: feynmanw@xjtu.edu.cn) are with the School of Software Engineering, Xi'an Jiaotong University, Xi'an 710049, China.}
\thanks{Yinghui He~(email: yinghui.he@ntu.edu.sg) is with the College of Computing and Data Science, Nanyang Technological University, Singapore 639798.}
\thanks{Han Ding~(email: dinghan@xjtu.edu.cn) and Ge Wang~(email: gewang@xjtu.edu.cn) are with the School of Computer Science and Technology, Xi'an Jiaotong University, Xi'an 710049, China. }
\thanks{Jinsong Han~(email: hanjinsong@xjtu.edu.cn) is with the College of Computer Science and Technology, Zhejiang University, Hangzhou 310027, China.}
\thanks{Fei Wang is the corresponding author.}
}

%
%

\markboth{Journal of \LaTeX\ Class Files,~Vol.~00, No.~0, January~2026}%
{MobiDiary: Autoregressive Human Action Captioning with Wearable Devices and Wireless Signals}
%



\maketitle

\begin{abstract}
Human Activity Recognition (HAR) in smart homes is critical for health monitoring and assistive living. While vision-based systems are common, they face privacy concerns and environmental limitations (e.g., occlusion). In this work, we present MobiDiary, a framework that generates natural language descriptions of daily activities directly from heterogeneous physical signals (specifically IMU and Wi-Fi). Unlike conventional approaches that restrict outputs to pre-defined labels, MobiDiary produces expressive, human-readable summaries. To bridge the semantic gap between continuous, noisy physical signals and discrete linguistic descriptions, we propose a unified sensor encoder. Instead of relying on modality-specific engineering, we exploit the shared inductive biases of motion-induced signals—where both inertial and wireless data reflect underlying kinematic dynamics. Specifically, our encoder utilizes a patch-based mechanism to capture local temporal correlations and integrates heterogeneous placement embedding to unify spatial contexts across different sensors. These unified signal tokens are then fed into a Transformer-based decoder, which employs an autoregressive mechanism to generate coherent action descriptions word-by-word. We comprehensively evaluate our approach on multiple public benchmarks (XRF V2, UWash, and WiFiTAD). Experimental results demonstrate that MobiDiary effectively generalizes across modalities, achieving state-of-the-art performance on captioning metrics (e.g., BLEU@4, CIDEr, RMC) and outperforming specialized baselines in continuous action understanding. 
\end{abstract}


%
\IEEEpeerreviewmaketitle

\section{Introduction}\label{sec:introduction}


The growing demand for intelligent, context-aware systems has driven major advances in Human Activity Recognition (HAR), particularly in smart homes. Monitoring daily activities supports healthcare, elder safety, and personalized assistance, while enabling more adaptive environments~\cite{kim2024health,li2025wilife,hong2025llm4har,bock2024wear,wang2024xrf55, xu2024penetrative}. For example, continuous monitoring can detect abnormal behaviors in the elderly—such as falls, inactivity, or cognitive decline—allowing timely intervention. For children, it enhances safety by identifying risky actions or restricted area entry. For patients in rehabilitation or with chronic conditions, motion tracking enables remote therapy and health monitoring. However, traditional HAR systems are often limited to classifying activities into discrete, predefined labels (e.g., walking, sitting). Human Action Captioning (HAC) advances this paradigm by instead generating continuous, natural language descriptions of user actions. HAC provides a far more expressive, interpretable, and human-readable summary of complex behaviors, offering richer context and a more intuitive understanding of daily routines than discrete labels can convey. Visual HAC, typically implemented via camera-based systems~\cite{venugopalan2015sequence, zhou2018end,shi2023learning, islam2024video}, has achieved high performance, but it raises serious privacy concerns in home scenarios. Moreover, achieving comprehensive coverage for robust tracking often requires the deployment of cameras in every room, which substantially increases 
cost and complexity.

\begin{figure*}[t]
    \centering
    \includegraphics[width=0.92\linewidth]{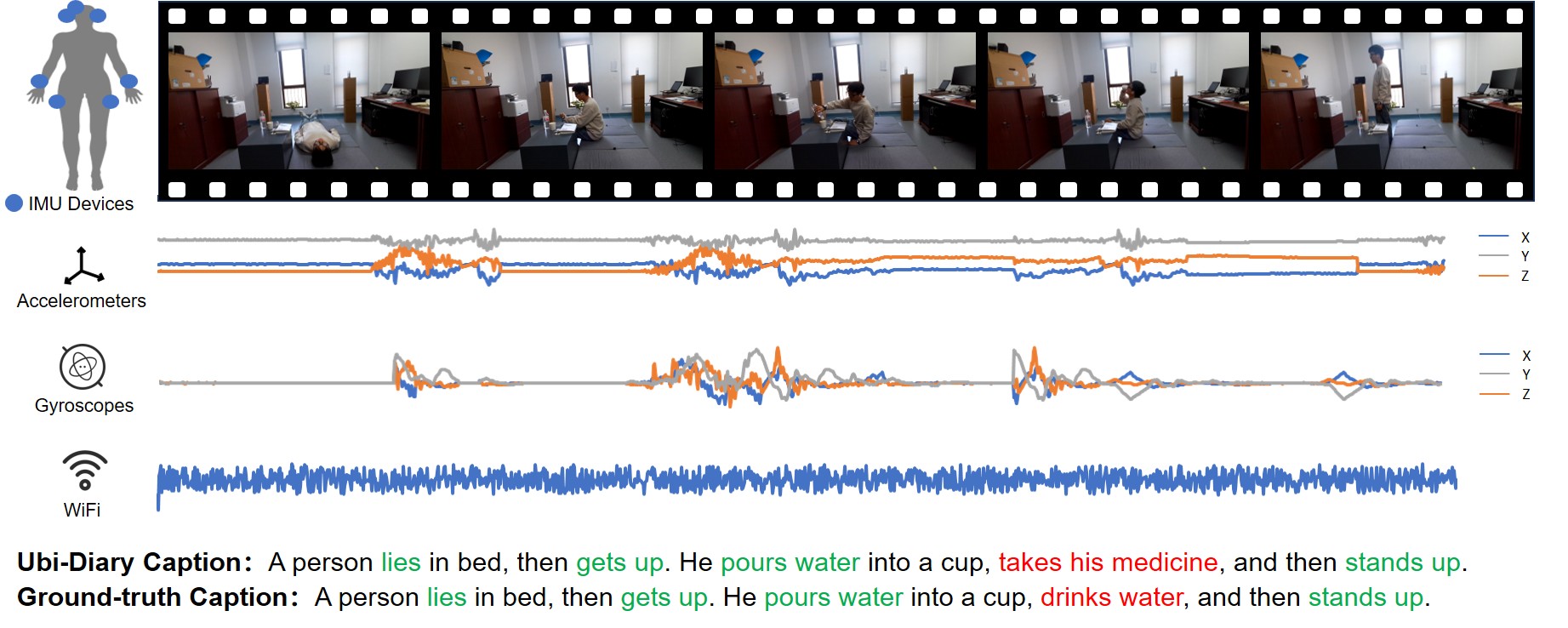}
    \vspace{-5pt} 
    \caption{A user moves around a home environment. \sysName{} captures the sensor signal sequence (originating from either worn IMU-equipped devices or ambient WiFi signals) and generates natural language descriptions of the user's activities. These textual summaries can serve as interpretable logs and support downstream tasks such as activity retrieval, monitoring, and assistive reasoning, especially when integrated with 
    LLMs to enhance contextual understanding and reasoning capabilities.}
    \label{fig:fig1}
    \vspace{-10pt} 
\end{figure*}

Wearable devices with inertial measurement units (IMUs) and wireless signals provide privacy-preserving and practical solutions, owing to their ubiquity in everyday devices such as smartphones, wearables, and home routers~\cite{bock2024wear,wang2015deepfi,zheng2019zero,
kotaru2015spotfi}. They are not only cost-effective and compact but also ensure reliable operation even under occluded or low-visibility conditions. However, most existing methods are designed for a single modality and categorize activities into coarse, predefined classes (e.g., walking, cooking), which fail to model the subtle variations that differentiate similar actions or capture the composite, multi-part nature of complex real-world activities. 
To address these limitations, we propose \texttt{\sysName}. As shown in Fig.~\ref{fig:fig1}, \texttt{\sysName} continuously processes IMUs or Wi-Fi signals collected from everyday devices, such as wearable devices and Wi-Fi adapters, and converts them into high-level semantic textual descriptions to build a comprehensive activity log. 
It is capable of richly expressing subtle motion details, and its natural language output is inherently compatible with large language models (LLMs), enabling it to serve as an intelligent agent in applications such as remote health monitoring and smart home automation.

To realize \texttt{\sysName}, two key challenges remain to be addressed. First, both IMU and Wi-Fi data are inherently low-level and lack explicit semantic structure, making it difficult to extract descriptive behavioral patterns. Second, these signals vary significantly depending on the sensor configuration (e.g., IMU placement or user location relative to Wi-Fi transceivers), which introduces spatial ambiguity that undermines generalization. To tackle these issues, we design a unified sensor encoder based on two principles: (1) enhancing local temporal semantic representations to identify action boundaries and capture discriminative motion patterns; and (2) embedding physical priors to improve spatial awareness and robustness to the diversity of sensor configurations.
Furthermore, drawing inspiration from the next-token prediction paradigm of LLMs, we design our model to be Autoregressive. This means it generates the action description word-by-word, conditioning the prediction of each new word on the sequence of words already generated, thus ensuring the construction of a contextually coherent text sequence.





To comprehensively evaluate the effectiveness of our approach, we conduct extensive experiments on four large-scale public datasets that cover both IMU and Wi-Fi modalities, including IMU of XRF V2 dataset~\cite{lan2025xrf}, IMU of UWash dataset~\cite{wang2025you}, Wi-Fi of XRF V2 dataset~\cite{lan2025xrf}, and Wi-Fi of WiFiTAD dataset~\cite{liu2025wifi}. These datasets feature continuous activity sequences from multiple participants, recorded across diverse indoor environments using various devices, and notably, cover distinct types of tasks, such as general daily activities and specialized handwashing actions. This diversity in both modalities and tasks is crucial for validating the generalizability of our method. Our results demonstrate that \texttt{\sysName} significantly outperforms existing methods across multiple captioning metrics. This includes classic video captioning models, such as RCG~\cite{zhang2021open} and VTAR~\cite{shi2023learning}, as well as advanced models designed for specific modalities like RF-Diary~\cite{fan2020home}, XRFMamba~\cite{lan2025xrf}, and WiFiTAD~\cite{liu2025wifi}, showcasing its formidable capability to understand and describe continuous signals from different modalities.


    
    
Our main contributions are as follows:
\vspace{-0.2\baselineskip}
\begin{itemize} 
    \item We propose \texttt{\sysName}, a novel general framework that generates natural language descriptions of daily activities directly from either raw IMU or Wi-Fi signals, bridging the gap between low-level sensor data and high-level semantic understanding.
    
    \item We design a unified sensor encoder that incorporates local temporal semantics and physical priors, enabling robust segmentation of continuous actions across diverse sensor modalities and configurations.
    
    \item We conduct comprehensive experiments on four multiple public IMU and Wi-Fi datasets, demonstrating that \texttt{\sysName} outperforms a wide range of state-of-the-art, specialized methods, thereby validating the superior performance and generalizability of our framework.
\end{itemize}

\section{Related Work}\label{sec:related_work}

\subsection{IMU and Wi-Fi Human Sensing}\label{sec:imu-sensing}
Inertial measurement units (IMUs), commonly embedded in wearable devices, have been widely used for human activity recognition due to their low cost, ubiquity, and ability to function without external infrastructure. Early methods use Convolutional Neural Networks and Long Short-Term Memory (LSTM) networks to model short-term or static actions from sequential IMU signals~\cite{anguita2013public}. To enhance robustness, MMG-Ego4D introduced modality dropout and contrastive learning for missing-modality scenarios~\cite{gong2023mmg}, while EgoPAT3D fused IMU and RGB-D data for 3D trajectory prediction in intent understanding tasks~\cite{li2022egopat3d}. SIP pioneered sparse-IMU-based 3D pose reconstruction~\cite{von2017sparse}, and PIP further improved stability using physical priors~\cite{yi2022physical}. With the emergence of large language models (LLMs), researchers have begun leveraging LLMs to augment IMU-based models. LLM4HAR \cite{hong2025llm4har} exploits the generalization prowess of LLMs to construct robust human activity recognition (HAR) frameworks; more aggressive approaches directly adopt LLMs as the inference backbone~\cite{ji2024hargpt, li2024sensorllm, imran2024llasa}; other efforts align IMU sequences with textual descriptions to inject semantic information, thereby improving cross-domain generalization~\cite{wang2025ego4o, zhang2024unimts}.

Wi-Fi-based human sensing leverages ubiquitous wireless infrastructure for device-free activity recognition and motion tracking in indoor environments~\cite{10556745}. The introduction of channel state information (CSI) was a key enabler, allowing for finer-grained perception by exploiting subcarrier-level amplitude and phase variations \cite{halperin2011tool,pu2013whole}. The advent of deep learning further enhanced CSI-based systems; early models like DeepFi~\cite{wang2015deepfi} improved robustness using autoencoders and phase calibration. This trend continued as more recent works incorporated sophisticated architectures like CNNs~\cite{wang2019person} and RNNs~\cite{zheng2019zero} to tackle complex tasks such as human pose estimation and action recognition. Beyond standard CSI analysis, advanced techniques also utilize Doppler profiles and Angle of Arrival (AoA) estimation to achieve a more precise localization~\cite{kotaru2015spotfi,wang2024multi}.

Despite these advancements, research on the semantic understanding of continuous actions remains limited, with most works concentrating on temporal action localization, such as XRFMamba~\cite{lan2025xrf}, WEAR~\cite{bock2024wear,bock2024temporal}, and WiFiTAD~\cite{liu2025wifi}. Till now, a pioneering RF-Diary~\cite{fan2020home} proposed a method for daily life captioning using privacy-preserving radio frequency (RF) signals combined with a home floormap. To overcome the lack of labeled RF data, it introduced a multi-modal feature alignment scheme that leverages large, existing video-captioning datasets to supervise the training of the RF-based model. Sensor2Text~\cite{chen2024sensor2text} extended this concept to enable natural language conversations (Q\&A) about daily activities using multi-modal wearable sensors. To handle low sensor information density, Sensor2Text employs a teacher-student network, using a pre-trained visual encoder as a ``teacher" to distill knowledge from paired video into the sensor encoder.

However, a significant limitation of both RF-Diary and Sensor2Text is their heavy reliance on video data during training. RF-Diary requires video datasets for its feature alignment, and Sensor2Text explicitly uses a visual teacher model, which necessitates paired sensor-video data for its cross-modal training phase. This dependency on video supervision restricts their applicability in broader scenarios where collecting paired video is impractical or violates the core privacy motivations for using non-visual sensors. In contrast, our method is designed to operate directly from IMU or Wi-Fi signals without requiring any video data for supervision.

\subsection{Human Action Captioning}\label{sec:action-captioning}



Video captioning is the most extensively studied paradigm in action captioning, aiming to generate fluent and accurate natural-language descriptions for unlabeled video clips. It typically follows an encoder-decoder framework, where a visual encoder extracts spatio-temporal features and a language decoder autoregressively generates the target sentence. The most significant performance improvements in video captioning stem primarily from advancements in both the encoder and decoder~\cite{abdar2024review, liu2025survey, rafiq2023video}.
Early models employed 2D/3D CNNs to extract visual features from frames or clips, paired with LSTMs for text generation~\cite{venugopalan2015sequence, yao2015describing, pan2016hierarchical}. Venugopalan \etal~\cite{venugopalan2015sequence} proposed an end-to-end sequence-to-sequence approach that maps mean-pooled CNN features directly to sentences. Subsequent works introduced attention-based LSTMs, dynamically attending to relevant temporal regions to substantially enhance description quality.

In recent years, Transformer-based encoders and decoders have become dominant due to their superior ability to model long-range dependencies \cite{lin2022swinbert}. Zhou \etal~\cite{zhou2018end} introduced the first fully Transformer-based end-to-end model, achieving state-of-the-art results on standard benchmarks. Moreover, pre-trained vision-language models have been widely adopted \cite{tewel2022zero, seo2022end, li2023lavender}, with CLIP~\cite{radford2021learning} serving as a prominent example. As a powerful encoder, it significantly enhances cross-modal feature alignment \cite{yang2021clip, tang2021clip4caption, tang2021clip4caption++}.
Although numerous studies have attempted to incorporate supplementary information (e.g., object relations or scene context) to further improve performance, most such approaches yield only marginal gains. This motivates us to shift our focus to encoder design, particularly given that Transformer-based text decoders and generation paradigms have become well-established \cite{zhang2019object, wang2019controllable, chen2019motion, zhang2020object, zheng2020syntax, pan2020spatio}.

\section{The Design of \sysName}\label{sec:Method}

\begin{figure*}[t]
    \centering
    \includegraphics[width=0.95\linewidth]{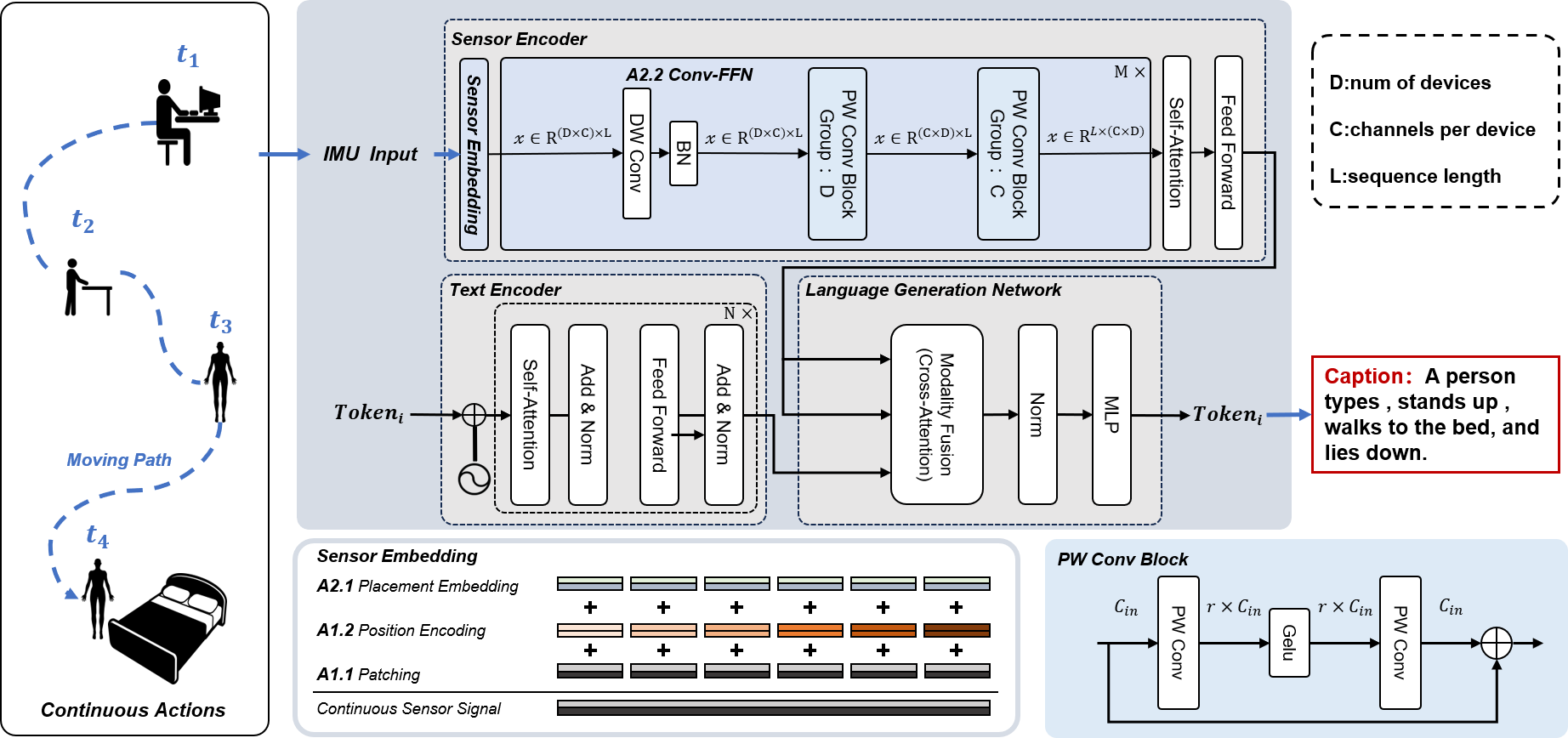}
    \caption{Overview of \sysName{}.
    During training, the input IMU or Wi-Fi sequence is encoded by the Sensor Encoder to extract motion features, while the ground-truth text description is processed by a text encoder. The fused features are then fed into a Language Generation Network to produce the action caption. During inference, the Text Encoder is initialized with a start-of-sequence token (\texttt{<sos>}), and action captions are generated in an Autoregressive manner through next-word prediction.}
    \label{fig:IMU-Diary-Model-Overview}
    \vspace{-5pt}
\end{figure*}

\subsection{Overview}\label{sec:overview}





Fig.~\ref{fig:IMU-Diary-Model-Overview} illustrates the overall architecture of \sysName, which is designed to process two distinct signal modalities: IMU and Wi-Fi. As a user moves within an indoor environment, the system records either motion data from wearable IMU devices or CSI derived from Wi-Fi signals. These raw sensor signals are first fed into a modality-specific \textbf{Sensor Encoder} to extract high-level temporal features.

During training, each sensor sequence is paired with a corresponding natural language description. The textual description is processed by a \textbf{Text Encoder}, after which the encoded textual features interact with the modality-specific features extracted from the sensor signals via a cross-attention mechanism. The resulting fused representation is then passed to a \textbf{Language Generation Network} to produce a natural language description of the activity. Further details on the training strategy are provided later in this section.

During inference, \sysName employs an Autoregressive generation strategy. Specifically, the input sensor sequence (either IMU or Wi-Fi) and a special \texttt{<SOS>} (start-of-sentence) token are fed into their respective sensor and text encoders. The model predicts the first token based on these inputs. The newly generated token is then concatenated with the \texttt{<SOS>} token and re-encoded by the text encoder to predict the next token. This process recurses until the \texttt{<EOS>} (end-of-sentence) token is generated, yielding a complete natural language description of the input sensor data.

In the following subsections, we elaborate on the key design details and underlying principles of each module in \sysName.


\subsection{Unified Sensor Encoder}\label{sec:imu-encoder}




Before detailing the architecture, we first address a fundamental question: Why can distinct modalities like IMU and Wi-Fi be processed by a unified encoder? Although they originate from different physical mediums, i.e., inertial measurements versus electromagnetic waves, they share fundamental inductive biases rooted in their generation mechanism
\begin{description}
 \item[Common Kinematic Causality] 
Both modalities are essentially time-series projections of the same underlying human motor behavior. An action (e.g., ``standing up") generates a unique kinematic signature that simultaneously manifests as a surge in accelerometer readings and a fluctuation in Wi-Fi Channel State Information (CSI). Thus, the latent semantic information is invariant across modalities.
\item[Local Signal Stationarity] 
Both signals exhibit high sampling rates relative to human motion speed. Consequently, adjacent data points possess high redundancy and strong local correlations. This shared characteristic justifies a unified Patching strategy to aggregate local frames into semantic tokens.
\item[Spatial Sensitivity] 
Both signals are heavily dependent on the spatial configuration (e.g., sensor placement or Tx-Rx location), necessitating a unified mechanism to embed Physical Priors (Placement Embeddings) to disentangle motion patterns from environmental noise.

\end{description}





Guided by these shared principles, we design the sensor encoder to address two specific challenges:


\begin{description}
    \item[Q1: Semantic Sparsity of Sensor Signals] 
    Both IMU and Wi-Fi signals are inherently low-level physical measurements that lack explicit semantic structure. IMU data reflects raw acceleration and angular velocity, while Wi-Fi signals (such as CSI) represent signal propagation characteristics. Extracting high-level patterns descriptive of complex human behaviors from such low-level data presents a significant challenge.

    \item[Q2: Signal Heterogeneity from Deployment Context] 
    The patterns in sensor signals are highly sensitive to their physical deployment context, which poses a severe test to model generalization.
    \begin{itemize}
        \item For \textbf{IMU signals}, patterns vary dramatically with the specific on-body placement of the device (e.g., wrist vs. pocket).
        \item Similarly, for \textbf{Wi-Fi signals}, patterns are extremely sensitive to the hardware properties and spatial layout of the transmitter-receiver (Tx-Rx) pair. Using different models of routers and network interface cards, or merely altering the deployment locations, can result in vastly different signal features for the exact same human activity.
    \end{itemize}
    This dependency on the deployment context makes it difficult for a model to generalize knowledge learned in one specific setting (e.g., User A with a phone in their pocket using a TP-Link router) to another (e.g., User B with a watch on their wrist using a Netgear router).
\end{description}
We design two complementary solutions for each challenge, namely A1.1 Patching and A1.2 Sinusoidal Position Encoding for Challenge 1, and A2.1 Placement Embedding and A2.2 Conv-FFN Block for Challenge 2, described next.

\subsubsection{A1.1 Patching} 


Enhancing local semantic representations is expected to facilitate the recognition of short-term motion primitives, such as hand-raising, turning, and walking, thereby aiding the model in delineating action boundaries and extracting discriminative motion patterns.
To this end, we adopt a patching strategy on the IMU or Wi-Fi signal, wherein temporally adjacent segments are grouped into single tokens to preserve and emphasize local temporal dependencies. This technique has been demonstrated to be effective in time series forecasting tasks~\cite{nie2022time}.

Concretely, let the patch length be denoted by \(P\), and the stride between two consecutive patches by \(S\). Given an input IMU sequence of length \(L\), the patching process produces a set of patches $\{x^{(i)}_p | i=1,2,..,N\} $, and the number of patches is computed as: $N = \lfloor \frac{L-P}{S} \rfloor + 1$. By integrating local information into each patch, the model is better equipped to capture fine-grained temporal relationships and encode meaningful local motion semantics.

\subsubsection{A1.2 Sinusoidal Position Encoding} 




Consistent with the design philosophy of A1.1, we employ modified sinusoidal positional encodings to further emphasize short-range dependencies and local motion patterns.
Although sinusoidal positional encoding is no longer a mainstream approach for the text modality, its inherent inductive bias provides a unique advantage in capturing instantaneous and local motion dynamics. 
Given Sinusoidal positional encoding 
\begin{equation}
\begin{cases}
    \mathbf{p}_{k, 2i} = \sin\left(\frac{k}{10000^{2i/d}}\right) \\
    \mathbf{p}_{k, 2i+1} = \cos\left(\frac{k}{10000^{2i/d}}\right)
\end{cases},
\end{equation}
where $\mathbf{p}_{k, 2i}$ and $\mathbf{p}_{k, 2i+1}$ are the $2i$-th and $(2i+1)$-th components of the positional encoding vector for position $k$, respectively, and $d$ is the vector dimension, the encoding at position $m$ is given by 
\begin{equation} \label{eq:your_label_here}
\mathbf{p}_m = 
\begin{pmatrix}
    \cos m\theta_0 \\
    \sin m\theta_0 \\
    \cos m\theta_1 \\
    \sin m\theta_1 \\
    \vdots \\
    \cos m\theta_{d/2-1} \\
    \sin m\theta_{d/2-1}
\end{pmatrix}
\Rightarrow
\mathbf{p}_m = 
\begin{pmatrix}
    e^{im\theta_0} \\
    e^{im\theta_1} \\
    \vdots \\
    e^{im\theta_{d/2-1}}
\end{pmatrix},
\end{equation} where $\theta_i = 10000^{-2i/d}$ in original Transformer~\cite{vaswani2017attention}. Consequently, the inner product between the encodings at positions $m$ and $n$ is expressed as 
\begin{align}
\langle \mathbf{p}_m, \mathbf{p}_n \rangle &= \text{Re} \left[ e^{i(m-n)\theta_0} + \dots + e^{i(m-n)\theta_{d/2-1}} \right] \notag \\
&= \frac{d}{2} \cdot \text{Re} \left[ \sum_{i=0}^{d/2-1} e^{i(m-n)10000^{-i/(d/2)}} \frac{1}{d/2} \right] \notag \\
&\sim \frac{d}{2} \cdot \text{Re} \left[ \int_{0}^{1} e^{i(m-n) \cdot 10000^{-t}} dt \right].
\label{eq:sinusoidal_inner_product}
\end{align}

\begin{figure}[h]
    \centering
    \includegraphics[width=1\linewidth]{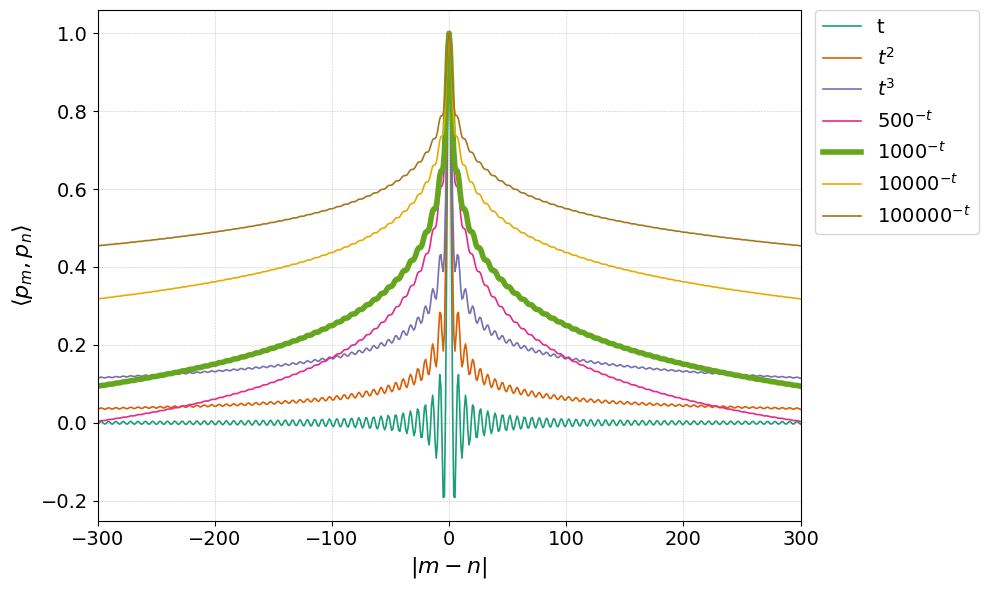}
    \caption{Sinusoidal Position Encoding. The decline of two positions' similarity of $\langle p_m, p_n \rangle$ as $|m-n|$ increases for different values of $\theta_t$.}
    \label{fig:curve}
    \vspace{-5pt}
\end{figure}

The problem thus reduces to the asymptotic estimation of the integral $\int_{0}^{1} e^{i(m-n)\theta_t} dt$. The value of the integral for different values of $\theta_t$ is shown in Fig.~\ref{fig:curve}. This shows that the similarity between two position encodings decreases as the distance \(|m - n|\) increases. When combined with the patching strategy, this implies that tokens representing temporally distant segments have lower similarity. The sinusoidal encoding thus naturally introduces a long-range decay property in token interactions, encouraging the model to focus more on nearby patches. This inductive bias is particularly well-suited for sensory data, where local temporal patterns are more informative than distant dependencies.


In our implementation, we set \(\theta_t = 1000^{-t}\), which leads to a smoother and faster decay of \(\langle p_m, p_n \rangle\), as shown in Fig.~\ref{fig:curve}. Empirically, this choice not only accelerates convergence during training but also results in improved model performance.

\subsubsection{A2.1 Plancement Embedding} 



We incorporate placement priors into the encoder design to enable the model to distinguish signals originating from different hardware devices. The core intuition is as follows:

For IMU signals, if the model knows whether a signal comes from a smartwatch on the wrist or a headset on the head, it can associate that signal with the corresponding motion patterns.
Similarly, for Wi-Fi signals, if the model knows that a signal is generated by a “TP-Link router and Intel NIC” pair, as opposed to a “Netgear router and a different NIC,” or a transceiver pair at a different location, it can associate the signal features with the specific hardware configuration, thereby effectively isolating the perturbations caused purely by human activity.

To embed this placement awareness, we introduce a trainable placement embedding for each signal source. For an IMU device, this embedding represents its on-body placement; whereas for Wi-Fi, it represents a specific Tx-Rx pair. Each source ID (whether an IMU placement ID or a Wi-Fi link ID) is mapped to a learnable vector via an embedding layer, which is then fused with the corresponding sensor signal.

This unified design provides the model with a structural hardware context, helping it to disentangle the intrinsic human activity patterns from device-specific signal variations, thereby greatly enriching the semantic representation of the sensor input and uniformly addressing Challenge Q2.

\subsubsection{A2.2 Conv-FFN Block}

Each IMU device captures six channels of motion data, including 3-axis accelerometer and 3-axis gyroscope signals. To enhance the model's understanding of the relationships across different IMU devices and sensor channels, we incorporate a convolutional feed-forward network (Conv-FFN) block~\cite{luo2024moderntcn}, as shown in Fig.~\ref{fig:IMU-Diary-Model-Overview}. This module adopts the core idea of Depthwise Separable Convolution \cite{howard2017mobilenets}, a computationally efficient alternative to standard convolution that consists of DWConv (Depth-Wise Convolution) and PWConv (Point-Wise Convolution). In \sysName, this structure is leveraged to efficiently model the complex dependencies both between channels (inter-channel) and between devices (inter-device).

In \sysName, the processing steps within the Conv-FFN are as follows. Given an input IMU sequence \(x \in \mathbb{R}^{D \cdot C \cdot L}\), where \(D\) is the number of devices, \(C\) is the number of sensor channels per device, and \(L\) is the sequence length, we first merge the device and channel dimensions, reshaping the input to \(x \in \mathbb{R}^{(D \cdot C) \cdot L}\). This transformation is aimed at reducing computational complexity and enabling the reshaped sequence to be processed by DWConv, allowing for unified data processing across devices and channels. This facilitates the effective extraction of temporal features and allows for independent modeling of temporal features for each channel. 

Next, a PWConv Block is applied with the number of groups set to \(D\), so that every group of \(C\) channels corresponds to all sensor channels from a single IMU device. 
This enables effective inter-channel modeling within each device. 
Then, another PWConv Block is applied to the output of the first PWConv  Block \(x \in \mathbb{R}^{(C \cdot D) \cdot L}\), capturing cross-device dependencies by modeling interactions across the same channel (e.g., x-acceleration) from different devices. As a result, the model can independently capture inter-channel dependencies (e.g., between acceleration and rotation) and inter-device relationships (e.g., hand vs. head movement).

Beyond IMU data, the Conv-FFN module is effectively adapted to process Wi-Fi signals by analogizing a Tx-Rx link to a ``device" (D) and a CSI subcarrier to a ``channel" (C). Following an initial DWConv that extracts temporal features for each subcarrier independently, the module employs a two-stage PWConv Blocks. First, a grouped convolution (groups=D) models cross-subcarrier dependencies, capturing the characteristic signature across the subcarriers within each single spatial path. Subsequently, two standard PWConv Blocks fuse these features to model cross-link dependencies, thereby integrating spatial information from multiple signal paths to form a global representation of the activity. This hierarchical approach enables the structured extraction of robust features from high-dimensional CSI data by systematically modeling temporal, spectral (cross-subcarrier), and spatial (cross-link) relationships.

As a complement, we introduce a self-attention module at the end of the Sensor Encoder to capture potential long-range dependencies. The Conv-FFN and self-attention modules offer complementary modeling capabilities, jointly enhancing the representation of IMU or Wi-Fi time-series signals.

\subsection{Text Encoder}

The second component of \sysName is the Text Encoder, which is responsible for encoding the natural language descriptions associated with the sensor signals into high-level feature representations. We adopt a standard Transformer encoder architecture \cite{zhou2018end}. Prior to being processed, the input text is first tokenized and converted into a sequence of vectors via a learnable embedding layer. As the Transformer architecture itself does not process sequential order, we inject positional encodings into the input embeddings to ensure the model can leverage word order information.

The resulting sequence of position-aware embeddings is then fed into an encoder composed of a stack of $N$ identical layers. Each layer consists of two core sub-modules: a multi-head self-attention network, which captures intra-text dependencies, and a position-wise feed-forward network. To ensure stable and effective training, each sub-module is wrapped with a residual connection followed by Layer Normalization.

\subsection{Language Generation Network}




The final component of \texttt{\sysName{}} is the Language Generation Network, whose core task is to synthesize information from both the upstream sensor encoder and the Text Encoder to generate an accurate and coherent natural language description. This network takes two distinct feature sequences as input: the high-level temporal features \(x \in \mathbb{R}^{L_1 \times d}\) produced by the sensor encoder (either IMU or Wi-Fi), and the contextual text features \(t \in \mathbb{R}^{L_2 \times d}\) output by the Text Encoder, where \(L_1\) and \(L_2\) are the respective sequence lengths and \(d\) represents the hidden dimension of the model.

To effectively integrate these two modalities, we employ a cross-attention mechanism, as illustrated in Fig.~\ref{fig:IMU-Diary-Model-Overview}. This mechanism allows the model to dynamically focus on the most relevant parts of the sensor sequence while generating each word. In this configuration, the textual features \(t\) serve as the Query, which probes the Key and Value pairs derived from the sensor features \(x\). Intuitively, this design allows the model to ask, based on the text generated so far (the query), ``Which segment of the temporal sensor data is most important for predicting the next word?"

This process is formalized as follows. First, the input features are projected into query, key, and value matrices using three independent, learnable projection matrices \(\mathbf{W}^Q, \mathbf{W}^K, \mathbf{W}^V \in \mathbb{R}^{d \times d_k}\):
\begin{equation}
    \mathbf{Q} = t \cdot \mathbf{W}^Q, \quad
    \mathbf{K} = x \cdot \mathbf{W}^K, \quad
    \mathbf{V} = x \cdot \mathbf{W}^V.
\end{equation}
The fused representation, \(z\), is then computed using scaled dot-product attention:
\begin{equation}
    z = \text{softmax}\left( \frac{\mathbf{Q} \cdot \mathbf{K}^\top}{\sqrt{d_k}} \right) \cdot \mathbf{V}.
\end{equation}
The resulting tensor \(z \in \mathbb{R}^{L_2 \times d}\) is a deeply integrated representation where each token's embedding is enriched not only with its original textual context but also with the most pertinent temporal information from the sensor data.

To ensure stable and efficient training, these fused features are subsequently passed through a Layer Normalization step. This standard technique helps mitigate issues such as vanishing or exploding gradients and facilitates faster convergence. Finally, the normalized output is fed into a lightweight multi-layer perceptron (MLP), which serves as the final decoder. The role of this MLP is to project the rich, fused representation into a probability distribution over the entire vocabulary, thereby predicting the most likely next token in the sequence.

\subsection{Training and Inference Strategy}\label{sec:training-and-inference}

\begin{figure}[htbp]
  \centering
  \begin{subfigure}{0.48\linewidth}
    \centering
    \includegraphics[width=\linewidth]{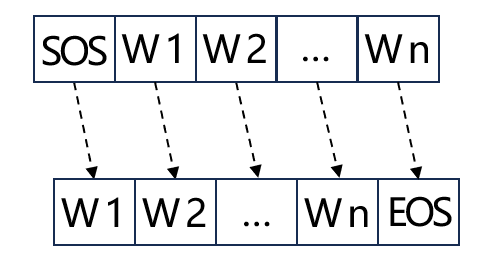}
    \caption{Teacher Forcing}
    \label{fig:teacher-forcing}
  \end{subfigure}
  \begin{subfigure}{0.48\linewidth}
    \centering
    \includegraphics[width=\linewidth]{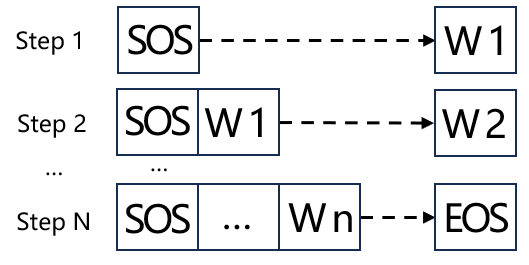}
    \caption{Autoregressive Generation}
    \label{fig:Autoregressive}
  \end{subfigure}
  \caption{During training, we employ the Teacher Forcing strategy, whereas during inference, we adopt an Autoregressive generation approach.}
  \label{fig:training-and-inference}
  \vspace{-5pt}
\end{figure}

\subsubsection{Training Strategy}
To achieve efficient and stable model convergence, we employ the Teacher Forcing training mechanism~\cite{vaswani2017attention}. At each decoding timestep, this strategy forcibly feeds the ground-truth token sequence as contextual input, effectively preventing the progressive accumulation of errors in the Autoregressive generation process, mitigating training instability caused by early prediction deviations, and significantly accelerating gradient convergence. The optimization objective adopts the standard Cross-Entropy Loss, formulated as:
\[
\mathcal{L} = -\sum_{t=1}^{T} \log P(w_t^* \mid w_{<t}^*, \mathbf{x}; \theta),
\]
where $w_t^*$ denotes the $t$-th ground-truth token, \textbf{$\mathbf{x}$} represents the encoded input IMU or Wi-Fi sequence, and $\theta$ denotes the model parameters.


To enhance the conciseness, semantic consistency, and generalization capability of the generated text, we strictly constrain the output vocabulary to the top 1000 high-frequency words from the training corpus (Vocabulary Size = 1000)~\footnote{\url{https://gist.github.com/deekayen/4148741}}. This design is motivated by the following considerations:  
(1) Information Concentration: High-frequency words cover the primary semantic expression needs, sufficient to support clear and structured action descriptions;  
(2) Overfitting Mitigation: Compressing the output space dimension reduces the model's memorization burden on low-frequency words;  
(3) Real-Time Efficiency: Reducing softmax computation overhead improves throughput during inference.  
This strategy ensures that generated descriptions are concise and robust while effectively alleviating the risk of overfitting.

\subsubsection{Inference Process}
The inference phase employs an Autoregressive decoding paradigm, formalized as:
\[
\hat{w}_t = \arg\max_{w \in \mathcal{V}} P(w \mid \hat{w}_{<t}, \mathbf{x}; \theta), \quad t = 1, 2, \dots,
\]
where $\mathcal{V}$ is the constrained vocabulary of 1000 words, and $\hat{w}_{<t}$ denotes the sequence of previously generated tokens. Decoding initiates from the special start token \texttt{<SOS>}, and the model sequentially generates the next token based on the global temporal features extracted by the encoder and the current context, until the end token \texttt{<EOS>} is produced or a predefined maximum length $T_{\max} = 50$ is reached.

\subsubsection{Computational Performance and Deployment Feasibility}


\sysName{} processes a 30-second, 100 Hz-sampled 6-DoF IMU sequence with an end-to-end latency of approximately 0.6 seconds on an NVIDIA RTX 3090 GPU (1.0 GB VRAM) and 5.0 seconds on an Intel i7-13700F CPU (1.2 GB RAM). For Wi-Fi CSI data, inference latency is slightly higher than the IMU modality due to the high-dimensional spatiotemporal features of CSI. On the same hardware, a 30-second Wi-Fi sequence incurs latencies of approximately 0.8 seconds (RTX 3090, 1.3 GB VRAM) and 6.5 seconds (i7-13700F, 1.5 GB RAM).
We therefore believe that for the human action caption generation task, a 5-second inference delay fully satisfies the requirements of non-real-time applications, such as offline analysis, rehabilitation logging, and daily behavior journaling.
\section{Experiments}\label{sec:experiments}

\subsection{Experimental Setup}\label{sec:setup}

\subsubsection{Dataset}

We consider four publicly available datasets for IMU-based action captioning: UWash~\cite{wang2025you}, Ego-ADL~\cite{sun2024multimodal}, XRF V2~\cite{lan2025xrf}, and WEAR~\cite{bock2024wear}. The Ego-ADL dataset contains IMU data from smartphones during users' daily activities. Although the dataset records continuous motion, its annotations are automatically generated using acoustic cues. Upon manual inspection, we find the label quality to be inconsistent and unreliable for our task. The WEAR dataset is designed for activity recognition in outdoor sports scenarios, with a focus on classifying the entire activity rather than generating detailed, sequential action descriptions, making it unsuitable for our task. In contrast, the UWash dataset consists of IMU recordings from 51 participants performing a handwashing routine following WHO guidelines. Since these recordings adhere to a standardized hand hygiene procedure, the dataset provides precise, fine-grained action labels with timestamps, making it well-suited for the task of action captioning. XRF V2~\cite{lan2025xrf} is a large-scale dataset containing continuous activity sequences recorded from 16 participants in indoor environments. Each participant wears four types of IMU-equipped devices: two smartphones (pocket), two smartwatches (wrist), earbuds (ears), and glasses (head). Each sequence consists of 7–10 actions, each lasting 5–20 seconds, with total sequence durations ranging from 50 to 80 seconds. In total, the dataset includes 853 sequences spanning approximately 16 hours of recordings, with 682 sequences used for training and 171 for testing. Importantly, each action is precisely annotated with start and end timestamps, making XRF V2 well-suited for our evaluation. For training and assessment, we segment each sequence into 30-second clips and assign a corresponding caption. As shown in Fig.~\ref{fig:fig1}, these captions are generated by concatenating the constituent actions within the clip with appropriate subjects and conjunctions to form a natural and coherent description.




In addition to IMU-based analysis, we also utilize two public datasets containing Wi-Fi CSI data for behavior perception and caption generation from Wi-Fi signals, i.e., XRF V2~\cite{lan2025xrf} and WiFiTAD~\cite{liu2025wifi}. XRF V2 is a multimodal dataset that, in addition to IMU data, also collects synchronized Wi-Fi signals, which are equally applicable to the action captioning task. Furthermore, to better analyze long-term, untrimmed signals, we incorporate WiFiTAD dataset specifically designed for Wi-Fi CSI-based temporal activity detection. This dataset contains 553 untrimmed Wi-Fi CSI samples collected in an indoor environment, with a total of 2,114 precisely annotated activity instances. The dataset covers seven daily activities, including walking, running, and jumping. Each activity instance is annotated with precise start and end timestamps and a category label, providing high-quality temporal information for our model training.

\subsubsection{Evaluation Metrics} 






To comprehensively evaluate the performance of our proposed model, we employ a suite of five metrics. Four of these are established metrics from the field of natural language generation, designed to assess the quality of the generated text from different perspectives. The fifth metric specifically measures the semantic understanding of the described action.

\begin{itemize}
    \item \textbf{BLEU@4 (B@4)}: This metric measures the precision of n-gram (up to 4-grams) overlap between the generated caption and reference captions. A high BLEU score indicates strong textual similarity and fluency.

    \item \textbf{METEOR (M)}: To provide a more robust evaluation than BLEU, METEOR computes a score based on the harmonic mean of unigram precision and recall, while also considering synonymy and stemming. This allows it to better capture semantic similarity at the word level.

    \item \textbf{ROUGE-L (R)}: This metric evaluates content coverage by measuring the recall of the Longest Common Subsequence (LCS) between the generated and reference texts. It is particularly effective at assessing how well the generated caption captures the essential information from the references.

    \item \textbf{CIDEr (C)}: Originally developed for image captioning, CIDEr evaluates consensus by weighting n-grams using Term Frequency-Inverse Document Frequency (TF-IDF). This makes it highly sensitive to informative words that are both frequent in reference captions for a specific activity and rare in the overall dataset, serving as a strong indicator of semantic relevance.
\end{itemize}


To evaluate the semantic accuracy of generated captions beyond surface-level text matching, we employ the \textbf{Response Meaning Consistency (RMC)} metric proposed in XRF V2~\cite{lan2025xrf}. RMC assesses a caption by using a Large Language Model (LLM) to answer a predefined set of questions, \(Q = \{q_1, \dots, q_{N_q}\}\), based solely on the caption's content. The LLM-generated answers are then compared against ground-truth answers derived from the source video to measure semantic fidelity. The RMC score for a generated caption, \(C_{\text{gen}}\), is calculated as:
\begin{equation}
\text{RMC}(C_{\text{gen}}) = \frac{1}{N_q} \sum_{i=1}^{N_q} \text{Sim}(\text{LLM}(C_{\text{gen}}, q_i), A_{\text{gt}_i}),
\end{equation}
where \(\text{LLM}(C_{\text{gen}}, q_i)\) is the LLM's answer to question \(q_i\) based on caption \(C_{\text{gen}}\); \(A_{\text{gt}_i}\) is the corresponding ground-truth answer; \(\text{Sim}(\cdot, \cdot)\) is a semantic similarity function, yielding 1 if the answers are consistent and 0 otherwise. This assessment is typically performed by human auditors to ensure accuracy.

A high RMC score signifies that the generated caption is not only linguistically coherent but also semantically sound, providing a robust measure of its functional correctness for downstream reasoning tasks.

\subsubsection{Implement Details}
The model is implemented in PyTorch, using the Adam optimizer with an initial learning rate of 1e-4 and a weight decay coefficient of 1e-3. The training was conducted using two NVIDIA 3090 GPUs, with a batch size of 16. The model was trained on the entire dataset for 100 epochs, taking approximately 10 hours to complete.

\subsection{Main Results}

\subsubsection{Experiments on the IMU datasets}
\begin{table*}[t]
\centering
\caption{Performance comparison on IMU-based datasets: XRF V2 (IMU) and UWash. Our method, \sysName, consistently outperforms all baselines across both datasets. The lower section presents the results of our ablation study.}
\begin{tabular}{l ccccc ccccc}
\toprule
& \multicolumn{5}{c}{XRF V2 (IMU) dataset} & \multicolumn{5}{c}{UWash dataset} \\
\cmidrule(lr){2-6} \cmidrule(lr){7-11}
Model & B@4$\uparrow$ & M$\uparrow$ & R$\uparrow$ & C$\uparrow$ & RMC$\uparrow$ & B@4$\uparrow$ & M$\uparrow$ & R$\uparrow$ & C$\uparrow$ & RMC$\uparrow$ \\
\midrule
LSTM \cite{venugopalan2015sequence} & 0.315 & 0.521 & 0.501 & 0.497 & 0.588 & 0.439 & 0.523 & 0.499 & 0.505 & 0.434 \\
Attention-LSTM \cite{yao2015describing} & 0.333 & 0.533 & 0.512 & 0.517 & 0.607 & 0.498 & 0.561 & 0.554 & 0.561 & 0.491\\
Transformer \cite{zhou2018end} & 0.430 & 0.583 & 0.566 & 0.572 & 0.684 & 0.589 & 0.668 & 0.637 & 0.631 & 0.566 \\
RCG \cite{zhang2021open} & 0.473 & 0.613 & 0.603 & 0.585 & 0.693 & 0.663 & 0.719 & 0.687 & 0.699 & 0.621 \\
VTAR \cite{shi2023learning} & 0.527 & 0.629 & 0.624 & 0.629 & 0.702 & 0.685 & 0.748 & 0.701 & 0.720 & 0.643 \\
Video Recap \cite{islam2024video} & 0.491 & 0.623 & 0.609 & 0.610 & 0.688 & 0.676 & 0.718 & 0.694 & 0.687 & 0.628 \\
XRFMamba \cite{lan2025xrf} & 0.680 & 0.747 & 0.734 & 0.725 & 0.739 & 0.868 & 0.858 & 0.809 & 0.846 & 0.788 \\
PatchTST \cite{nie2022time} & 0.688 & 0.782 & 0.773 & 0.767 & 0.791 & 0.834 & 0.883 & 0.861 & 0.883 & 0.815 \\
ModernTCN \cite{luo2024moderntcn} & 0.683 & 0.791 & 0.791 & 0.785 & 0.789 & 0.842 & 0.878 & 0.871 & 0.894 & 0.823 \\
TimeLLM \cite{jin2023time} & 0.694 & 0.799 & 0.798 & 0.788 & 0.793 & 0.851 & 0.885 & 0.877 & 0.899 & 0.830 \\
SensorLLM \cite{li2024sensorllm}  & 0.723 & 0.804 & 0.828 & 0.824 & 0.733 & 0.828 & 0.873 & 0.844 & 0.861 & 0.710 \\
LLaSA \cite{imran2024llasa} & 0.702 & 0.788 & 0.809 & 0.810 & 0.745 & 0.833 & 0.881 & 0.851 & 0.869 & 0.730 \\
\midrule
\textbf{\sysName (ours)} & \textbf{0.742} & \textbf{0.841} & \textbf{0.838} & \textbf{0.834} & \textbf{0.855} & \textbf{0.866} & \textbf{0.904} & \textbf{0.893} & \textbf{0.908} & \textbf{0.843} \\
w/o patch & 0.678 & 0.761 & 0.770 & 0.755 & 0.776 & 0.803 & 0.838 & 0.840 & 0.846 & 0.791 \\
w/o PE & 0.724 & 0.822 & 0.803 & 0.797 & 0.826 & 0.863 & 0.903 & 0.894 & 0.906 & 0.855 \\
w/o Conv-FFN & 0.705 & 0.804 & 0.791 & 0.781 & 0.814 & 0.854 & 0.891 & 0.878 & 0.898 & 0.810 \\
\bottomrule
\end{tabular}
\label{tab:imu-results}
\end{table*}






Table ~\ref{tab:imu-results} presents the detailed experimental results on the XRF V2 (IMU) and UWash datasets. As the inaugural study to generate action descriptions from IMU signals, we implemented a comprehensive suite of baseline models for comparative analysis. We began by implementing three standard sequential models: LSTM~\cite{venugopalan2015sequence}, Attention-LSTM~\cite{yao2015describing}, and Transformer~\cite{zhou2018end}. Although these models feature structural improvements, their performance on this task remains limited.

Subsequently, we adapted three prominent video captioning models: RCG~\cite{zhang2021open}, VTAR~\cite{shi2023learning}, and Video ReCap~\cite{islam2024video}. RCG reformulates video captioning as an ``open-book" task by first retrieving relevant sentences from a text corpus and then employing a copy-mechanism generator to dynamically create richer descriptions. VTAR utilizes a retrieval unit to obtain relevant text as a ``semantic anchor" and then employs an alignment unit to align video and text features in a shared space, thereby generating more accurate descriptions. Video ReCap introduces a recursive architecture that generates captions hierarchically, leveraging descriptions from previous levels to guide the generation for subsequent ones, enabling multi-level, multi-granularity captioning for hour-long videos. While these models excel in video captioning, their capacity to capture the temporal and semantic complexities inherent in IMU data is limited.

We also reimplemented XRFMamba~\cite{lan2025xrf}. In its original design, this model fuses IMU and Wi-Fi signals and is the first to use Mamba, a State Space Model, as its backbone to efficiently capture long-term dependencies for precise indoor continuous action localization. In our experiments, we trained XRFMamba using only IMU signals. Although this may limit its multimodal fusion advantages, the model still achieved competitive performance.

Furthermore, we evaluated three advanced time series models as IMU signal encoders. PatchTST~\cite{nie2022time} segments long time series into smaller ``patches" as input tokens for a Transformer and processes each variable independently, significantly improving efficiency and accuracy in forecasting tasks. ModernTCN~\cite{luo2024moderntcn} modernizes the traditional Temporal Convolutional Network (TCN) by incorporating designs from modern computer vision networks, such as large convolution kernels, to create a pure convolutional architecture that rivals Transformer and MLP-based models in both performance and efficiency. TIME-LLM~\cite{jin2023time} ``reprograms" time series data into text prototypes that an LLM (Llama-7B) can understand, using prompts to guide the process without altering the LLM's architecture, thereby leveraging its powerful forecasting capabilities. While these time series models demonstrated strong performance, significant disparities between IMU signals and traditional time series data prevented them from achieving optimal results.

Finally, we tested two novel methods that directly leverage LLMs to process IMU data. SensorLLM~\cite{li2024sensorllm} employs a two-stage framework built upon Llama-2-13b-chat that first automatically translates sensor signals into text describing their trends (e.g., rising, falling), enabling an LLM to understand the temporal characteristics of raw sensor data without manual annotation. LLaSA first generates large-scale narrative (SensorCaps) and instruction-following (OpenSQA) datasets using GPT, then uses this data to train a multimodal agent integrating an IMU encoder and an LLM (Vicuna-7b-1.5) for a deep understanding of human activities. Inspired by LLaSA~\cite{imran2024llasa}, we trained a similar multimodal agent on the XRF V2 (IMU) dataset. Our experiments revealed that such LLM-based approaches excel on metrics evaluating textual semantics but perform relatively poorly on the RMC metric, which assesses action understanding. This suggests that while LLMs can generate fluent language, they may have limitations in deeply comprehending the intrinsic physical properties of IMU signals.


Our proposed \sysName{} model demonstrates a comprehensive lead on both the XRF V2 (IMU) and UWash datasets. Notably, on the RMC metric, which evaluates action understanding capability, our model achieves scores of 0.855 and 0.843 on the two datasets, respectively. This robustly proves its exceptional ability and strong generalization in generating action descriptions that are both linguistically fluent and semantically precise.






The lower section of Table~\ref{tab:imu-results} presents the results of our ablation study, designed to validate the effectiveness of the model's key components.

\begin{itemize}
    \item \textbf{w/o patch}: Removing the patch module resulted in a significant performance drop across both datasets. Notably, on the XRF V2 (IMU) dataset, the RMC score decreased substantially from 0.855 to 0.776. This indicates that the patching mechanism is crucial for capturing local temporal features and understanding sensor data.

    \item \textbf{w/o PE}: The removal of the PE (placement embedding) module led to a performance decrease on the multi-device XRF V2 (IMU) dataset (RMC dropped from 0.855 to 0.826), but had almost no impact on the single-device UWash dataset. This clearly demonstrates that the PE module is effective for handling data from heterogeneous devices, but provides no benefit in a single-device scenario like UWash, where it is functionally irrelevant.

    \item \textbf{w/o Conv-FFN}: Similarly, removing the Conv-FFN module caused a significant performance degradation on both datasets, with the RMC score on the UWash dataset dropping from 0.843 to 0.810. This demonstrates that the module plays a core role in feature extraction and representation learning.
\end{itemize}

In summary, the results of the ablation study validate the soundness of our model's design, confirming that each component makes an indispensable contribution to the final outstanding performance.

\subsubsection{Experiments on the Wi-Fi datasets}



Compared to IMU data, Wi-Fi signals are more abstract and typically contain a higher degree of environmental noise. To evaluate model performance on such data, we conducted experiments on two datasets: XRF V2 Wi-Fi and WiFiTAD. Given the inherent difficulty of the Wi-Fi-based captioning task, we excluded earlier models like LSTM, Attention-LSTM, and Transformer. Instead, we included two more representative baselines: RF-Diary and WiFiTAD.

WiFiTAD proposes a dual pyramid network. One pyramid decomposes the Wi-Fi signal into high and low-frequency components to learn semantic features, while the other directly captures signal fluctuations in a learning-free manner. The features from both pyramids are fused via a cross-attention mechanism to achieve precise temporal action localization. In its original design, RF-Diary combines 3D human skeletons extracted from radio signals with a person-centric floormap to understand human actions and object interactions, leveraging a multi-modal feature alignment scheme to transfer knowledge from existing video datasets. As the datasets used in our evaluation do not provide human skeleton or floormap data, we retained only its core model architecture for our experiments.









\begin{table*}[t]
\centering
\caption{Performance comparison on WiFi-based datasets: XRF V2 (Wi-Fi) and WiFiTAD. These datasets test the model's ability to handle more abstract and noisy sensor data. The lower section includes the ablation study.}

\begin{tabular}{l ccccc ccccc}
\toprule
& \multicolumn{5}{c}{XRF V2 (Wi-Fi) dataset} & \multicolumn{5}{c}{WiFiTAD dataset} \\
\cmidrule(lr){2-6} \cmidrule(lr){7-11}
Model & B@4$\uparrow$ & M$\uparrow$ & R$\uparrow$ & C$\uparrow$ & RMC$\uparrow$ & B@4$\uparrow$ & M$\uparrow$ & R$\uparrow$ & C$\uparrow$ & RMC$\uparrow$ \\
\midrule
RCG \cite{zhang2021open}& 0.213 & 0.255 & 0.249 & 0.253 & 0.218 & 0.253 & 0.283 & 0.291 & 0.288 & 0.244 \\
VTAR \cite{shi2023learning} & 0.224 & 0.261 & 0.251 & 0.266 & 0.234 & 0.269 & 0.301 & 0.296 & 0.299 & 0.261 \\
Video Recap \cite{islam2024video} & 0.215 & 0.251 & 0.239 & 0.252 & 0.221 & 0.257 & 0.279 & 0.293 & 0.286 & 0.233 \\
RF-Diary  \cite{fan2020home} & 0.312 & 0.362 & 0.367 & 0.352 & 0.216 & 0.332 & 0.364 & 0.371 & 0.373 & 0.288 \\
XRFMamba \cite{lan2025xrf} & 0.422 & 0.446 & 0.457 & 0.453 & 0.370 & 0.498 & 0.543 & 0.533 & 0.559 & 0.504 \\
WiFiTAD \cite{liu2025wifi} & 0.429 & 0.448 & 0.463 & 0.458 & 0.384 & 0.503 & 0.558 & 0.549 & 0.571 & 0.517 \\
PatchTST \cite{nie2022time} & 0.421 & 0.441 & 0.451 & 0.442 & 0.377 & 0.463 & 0.502 & 0.497 & 0.526 & 0.469 \\
ModernTCN \cite{luo2024moderntcn} & 0.423 & 0.448 & 0.453 & 0.457 & 0.385 & 0.468 & 0.513 & 0.505 & 0.535 & 0.468 \\
TimeLLM \cite{jin2023time} & 0.429 & 0.452 & 0.461 & 0.462 & 0.394 & 0.477 & 0.521 & 0.518 & 0.545 & 0.471 \\
SensorLLM \cite{li2024sensorllm} & 0.392 & 0.423 & 0.433 & 0.419 & 0.310 & 0.442 & 0.482 & 0.466 & 0.502 & 0.388 \\
LLaSA \cite{imran2024llasa} & 0.397 & 0.431 & 0.429 & 0.421 & 0.322 & 0.449 & 0.461 & 0.459 & 0.504 & 0.394 \\
\midrule
\textbf{\sysName (ours)} & \textbf{0.454} & \textbf{0.483} & \textbf{0.492} & \textbf{0.481} & \textbf{0.410} & \textbf{0.513} & \textbf{0.583} & \textbf{0.588} & \textbf{0.576} & \textbf{0.522} \\
w/o patch & 0.412 & 0.441 & 0.453 & 0.447 & 0.373 & 0.473 & 0.541 & 0.553 & 0.532 & 0.499 \\
w/o PE & 0.441 & 0.468 & 0.478 & 0.463 & 0.390 & 0.493 & 0.563 & 0.576 & 0.541 & 0.509 \\
w/o Conv-FFN & 0.439 & 0.459 & 0.471 & 0.462 & 0.399 & 0.503 & 0.571 & 0.580 & 0.555 & 0.515 \\
\bottomrule
\end{tabular}
\label{tab:wifi-results}
\end{table*}

As shown in Table~\ref{tab:wifi-results}, our proposed \sysName  model achieves state-of-the-art performance on both the XRF V2 (Wi-Fi) and WiFiTAD datasets, demonstrating a comprehensive lead over all baseline models. This result underscores the model's robustness and strong generalization capabilities, as it excels not only with IMU data but also with more abstract and noisy Wi-Fi signals.

On the XRF V2 (Wi-Fi) dataset, our model obtains the highest scores across all five metrics: B@4$\uparrow$ (0.454), M$\uparrow$ (0.483), R$\uparrow$ (0.492), C$\uparrow$ (0.481), and RMC$\uparrow$ (0.410). Notably, our model's RMC score of 0.410 surpasses even specialized methods like WiFiTAD (0.384) and XRFmamba (0.370), indicating a superior ability to capture the semantic essence of the underlying actions.

This trend of superior performance continues on the WiFiTAD dataset, where \sysName once again leads in all metrics with scores of 0.513 (B@4$\uparrow$), 0.583 (M$\uparrow$), 0.588 (R$\uparrow$), 0.576 (C$\uparrow$), and 0.522 (RMC$\uparrow$). It is particularly noteworthy that our model outperforms the WiFiTAD baseline on its own namesake dataset, further validating the advanced design of our architecture.

The lower section of Table~\ref{tab:wifi-results} presents the results of our ablation study, which validates the effectiveness of our model's key components.

\begin{itemize}
    \item \textbf{w/o patch}: Removing the patching mechanism leads to a substantial drop in performance across both datasets. For instance, on the WiFiTAD dataset, the RMC score falls from 0.522 to 0.499. This highlights the critical role of the patching strategy in capturing local temporal patterns within the Wi-Fi signals.

    \item \textbf{w/o PE}: The removal of the placement embedding (PE) module results in a consistent performance decrease. On the XRF V2 (Wi-Fi) dataset, the RMC score drops from 0.410 to 0.390. This suggests that the PE module effectively learns and compensates for device- or environment-specific variations, even in scenarios primarily focused on Wi-Fi.

    \item \textbf{w/o Conv-FFN}: Removing the Conv-FFN module causes a significant degradation in performance on both datasets. For example, on the WiFiTAD dataset, the RMC score declines from 0.522 to 0.515. This result confirms that the Conv-FFN module is a core component for feature extraction and representation learning in our architecture.
\end{itemize}

In summary, the ablation experiments confirm the soundness of our model's design, demonstrating that each component makes an indispensable contribution to its state-of-the-art performance.

\subsubsection{Qualitative Captioning Examples}

\begin{figure*}[h]
    \centering
    \includegraphics[width=0.98\linewidth]{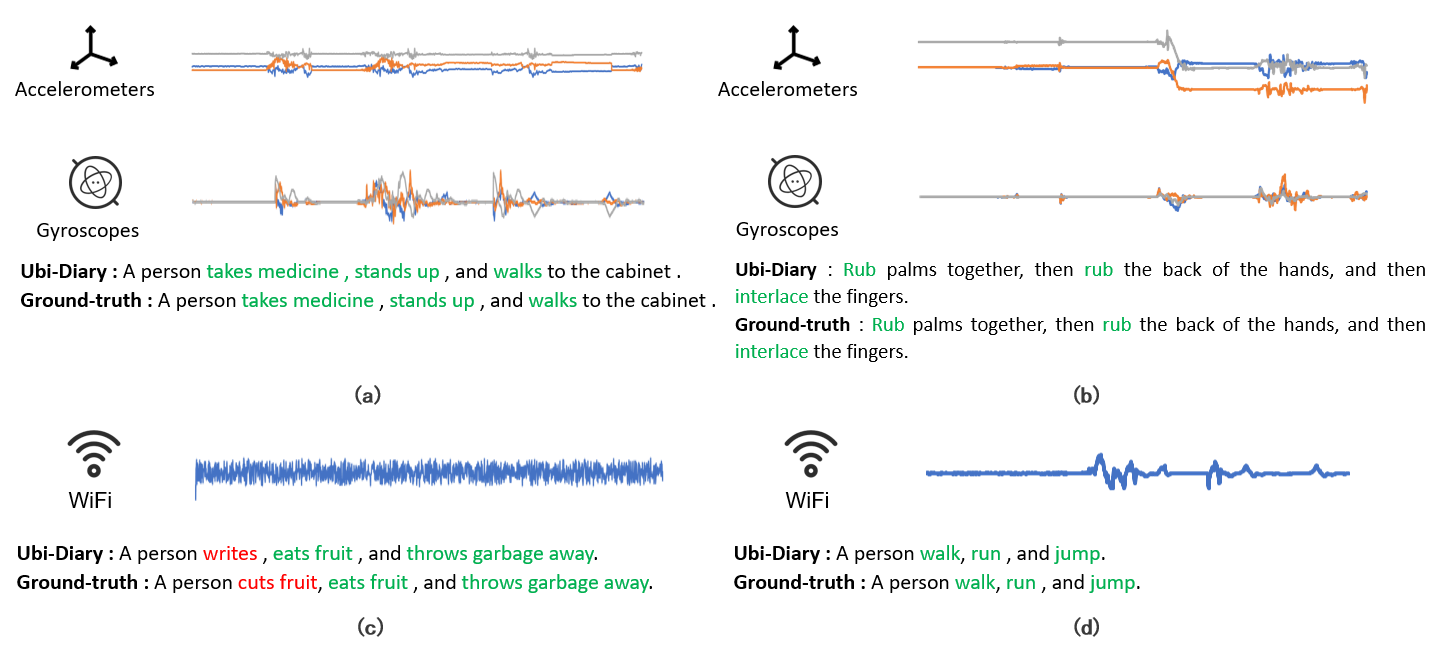}
    \caption{Qualitative results of the \sysName model on IMU (a, b) and WiFi (c, d) datasets. The model generated accurate descriptions on the XRF V2 (IMU) (a), UWash (b), and WiFiTAD (d) datasets, but struggled with abstract signals in the XRF V2 (WiFi) (c) dataset, confusing ``cuts fruit" with ``writes".}
    \label{fig:placeholder}
\end{figure*}

Fig.~\ref{fig:placeholder} provides qualitative examples of \sysName's performance on both IMU (a, b) and Wi-Fi (c, d) signals from our test sets, where keywords that match the ground truth are highlighted in green, while mismatches are shown in red. The figure illustrates the model's effectiveness across various scenarios. For instance, on the XRF V2 (IMU) (a), UWash (b), and WiFiTAD (d) datasets, the model generates descriptions that are largely consistent with the ground truth, successfully capturing the complex action sequences. Notably in (b), the model accurately identifies the fine-grained action ``interlace the fingers", showcasing its strong capability for detail recognition. However, the figure also reveals the model's limitations when processing particularly challenging signals. In (c), when faced with the noisy XRF V2 (Wi-Fi) signal, the model exhibits semantic confusion, misidentifying ``cuts fruit" as ``writes". This may be attributed to the similar small-scale hand movement patterns that both actions present in the Wi-Fi signals.

\subsection{Device Combination Study}

\begin{figure}[t]
    \centering
    \begin{minipage}{0.2\linewidth} 
        \centering
        \includegraphics[width=\linewidth]{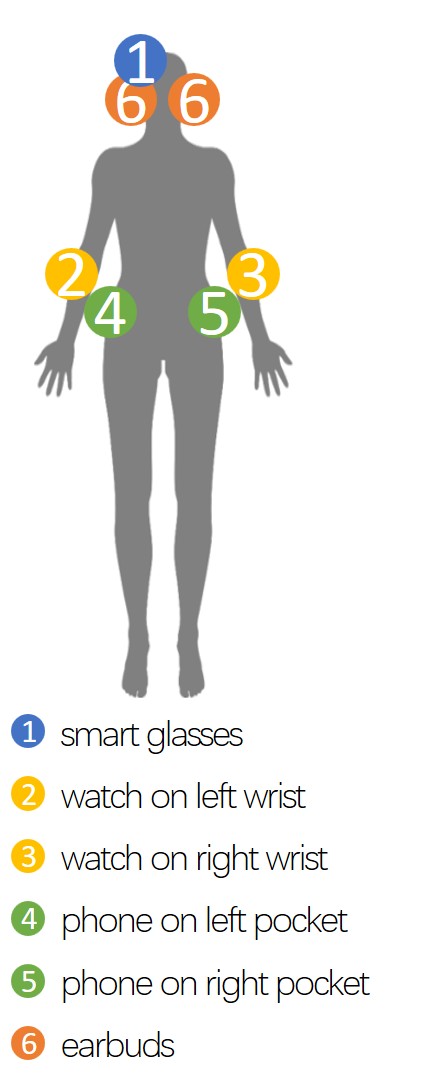}
    \end{minipage}%
    \hspace{20pt}
    \begin{minipage}{0.6\linewidth} 
        \small
        \setlength{\tabcolsep}{2pt}
        \resizebox{\linewidth}{!}{  
        \begin{tabular}{c c c c c c c}
        \toprule
        No.& Combination & B@4$\uparrow$ & R$\uparrow$ & M$\uparrow$ & C$\uparrow$ & RMC $\uparrow$ \\
        \midrule
        1 & 3 & 0.195 & 0.452 & 0.427 & 0.391 & 0.227 \\
        2 & 4 & 0.237 & 0.487 & 0.460 & 0.437 & 0.339 \\
        3 & 1\textbackslash3 & 0.307 & 0.542 & 0.515 & 0.490 & 0.432 \\
        4 & 2\textbackslash4 & 0.355 & 0.587 & 0.559 & 0.534 & 0.641 \\
        5 & 3\textbackslash4 & 0.412 & 0.618 & 0.591 & 0.580 & 0.699 \\
        6 & 2\textbackslash4\textbackslash5 & 0.445 & 0.646 & 0.621 & 0.610 & 0.711 \\
        7 & 1\textbackslash2\textbackslash4 & 0.450 & 0.683 & 0.662 & 0.648 & 0.729 \\
        8 & 1\textbackslash3\textbackslash5 & 0.495 & 0.672 & 0.652 & 0.646 & 0.725 \\
        9 & 1\textbackslash2\textbackslash4\textbackslash6 & 0.528 & 0.699 & 0.693 & 0.690 & 0.764 \\
        10 & 1\textbackslash3\textbackslash5\textbackslash6 & 0.595 & 0.672 & 0.651 & 0.645 & 0.783 \\
        11 & 1\textbackslash2\textbackslash3\textbackslash4\textbackslash6 & \underline{0.666} & \underline{0.791} & \underline{0.778} & \underline{0.773} & \underline{0.801} \\
        12 & 1\textbackslash2\textbackslash3\textbackslash4\textbackslash5\textbackslash6 & \textbf{0.742} & \textbf{0.841} & \textbf{0.838} & \textbf{0.819} & \textbf{0.855}\\
        \bottomrule
        \end{tabular}
        } 
    \end{minipage}
    \caption{We evaluate 12 typical device combinations from the XRF V2 (IMU) dataset. Results show that increasing the number of IMU devices generally improves performance.}
    \label{fig:devices-combination}
    \vspace{-5pt}
\end{figure}




The XRF V2 dataset comprises six types of IMU-equipped devices: smart glasses, two smartwatches (worn on the left and right wrists), two smartphones (carried in the left and right pockets), and a pair of earbuds, as illustrated in Fig.~\ref{fig:devices-combination}. Users can flexibly combine these devices based on their individual preferences or daily usage habits when using \sysName.





We evaluate several representative device configurations, with results presented in Table~\ref{fig:devices-combination}. Several key findings emerge from these results:
\begin{itemize}\vspace{-0.5\baselineskip} %
    \item \textbf{Importance of Device Type:} The single-device setups \textbf{(Cases 1--2)} establish a performance baseline. Notably, a wrist-worn watch \textbf{(Case 2)} significantly outperforms a phone in a pocket \textbf{(Case 1)}, suggesting that capturing direct limb-end motion (e.g., from the wrist) is more informative for generating descriptions than capturing indirect torso movement.
    \item \textbf{Strategic Placement Over Sheer Quantity:} The multi-device setups \textbf{(Cases 3--12)} show a clear upward performance trend. A core finding is that the strategic placement of sensors is more critical than merely increasing the device count. For instance, providing \textbf{bilateral body coverage} as in \textbf{Case 5} (watch on right wrist, phone in left pocket) significantly outperforms a unilateral setup like \textbf{Case 4} (watch and phone on the left side) across all metrics (e.g., 0.412 vs. 0.355 in B@4). This underscores the value of capturing a more holistic, full-body representation of the user's motion.
    \item \textbf{Value of Multi-Segment Body Coverage:} The top-performing configurations (e.g., \textbf{Cases 9 and above}) consistently involve coverage of multiple body segments (head, wrists, and torso). This enables the model to learn and correlate the synergistic movements between different body parts—for instance, associating the arm swing from a watch with the torso rotation from a phone—leading to richer and more accurate descriptions.
    \item \textbf{Diminishing Returns and Fusion Complexity:} As the device count increases to four or more, performance improvements exhibit diminishing returns. Interestingly, the transition from \textbf{Case 8} to \textbf{Case 10} (which adds a pair of earbuds) leads to a significant boost in the B@4 score but a slight degradation in the R, M, and C metrics. This reveals the complexity of multi-sensor fusion: not all additional information is purely beneficial, as it can sometimes introduce redundant or even conflicting signals. This provides an avenue for future research into more advanced fusion strategies.
\end{itemize}\vspace{-0.5\baselineskip} %

Importantly, we observe that \sysName achieves highly competitive results even with sparse, realistic setups. For example, with just one smartwatch and one smartphone in a bilaterally distributed configuration \textbf{(Case 5)}, the model achieves an R score of 0.618 and a METEOR score of 0.591.  This highlights the practical viability of \sysName, delivering robust performance even under realistic deployment constraints.

\subsection{Generalization to Extended Sequence Durations}

\begin{table*}
\centering
\caption{Performance Comparison Across Varying Input Sequence Lengths.}

\label{tab:long-sequence-ablation}
\begin{tabular}{@{} l l c c c c c c c c @{}}
\toprule
\textbf{Model} & \textbf{Length (s)} & \multicolumn{2}{c}{\textbf{XRF V2 (IMU)}} & \multicolumn{2}{c}{\textbf{UWash}} & \multicolumn{2}{c}{\textbf{XRF V2 (Wi-Fi)}} & \multicolumn{2}{c}{\textbf{WiFiTAD}} \\
\cmidrule(lr){3-4} \cmidrule(lr){5-6} \cmidrule(lr){7-8} \cmidrule(lr){9-10}
& & S-Avg$\uparrow$ & RMC$\uparrow$ & S-Avg$\uparrow$ & RMC$\uparrow$ & S-Avg$\uparrow$ & RMC$\uparrow$ & S-Avg$\uparrow$ & RMC$\uparrow$ \\
\midrule
 & 30 & 0.602 & 0.702 & 0.713 & 0.604 & 0.251 & 0.234 & 0.291 & 0.261 \\
& 45 & 0.608 & 0.709 & 0.717 & 0.610 & 0.250 & 0.233 & 0.290 & 0.260 \\
VTAR \cite{zhang2021open} & 60 & 0.592 & 0.693 & 0.705 & 0.595 & 0.228 & 0.210 & 0.268 & 0.240 \\
& 90 & 0.565 & 0.669 & 0.677 & 0.570 & 0.198 & 0.180 & 0.238 & 0.209 \\
& 120 & 0.535 & 0.638 & 0.645 & 0.538 & 0.155 & 0.140 & 0.195 & 0.165 \\
\midrule
 & 30 & 0.753 & 0.791 & 0.865 & 0.815 & 0.439 & 0.377 & 0.497 & 0.469 \\
& 45 & 0.758 & 0.795 & 0.871 & 0.819 & 0.436 & 0.374 & 0.493 & 0.465 \\
PatchTST \cite{nie2022time} & 60 & 0.742 & 0.778 & 0.852 & 0.803 & 0.418 & 0.355 & 0.478 & 0.450 \\
& 90 & 0.715 & 0.751 & 0.825 & 0.776 & 0.380 & 0.315 & 0.440 & 0.412 \\
& 120 & 0.680 & 0.715 & 0.785 & 0.735 & 0.330 & 0.265 & 0.385 & 0.350 \\
\midrule
 & 30 & 0.770 & 0.793 & 0.878 & 0.830 & 0.451 & 0.394 & 0.512 & 0.471 \\
& 45 & 0.775 & 0.798 & 0.882 & 0.835 & 0.447 & 0.390 & 0.508 & 0.467 \\
TimeLLM \cite{jin2023time} & 60 & 0.760 & 0.783 & 0.867 & 0.819 & 0.432 & 0.375 & 0.493 & 0.452 \\
& 90 & 0.735 & 0.758 & 0.840 & 0.793 & 0.395 & 0.338 & 0.455 & 0.415 \\
& 120 & 0.695 & 0.720 & 0.795 & 0.748 & 0.345 & 0.288 & 0.405 & 0.365 \\
\midrule
 & 30 & 0.795 & 0.733 & 0.852 & 0.710 & 0.417 & 0.310 & 0.473 & 0.388 \\
& 45 & 0.799 & 0.737 & 0.856 & 0.714 & 0.413 & 0.306 & 0.469 & 0.384 \\
SensorLLM \cite{li2024sensorllm} & 60 & 0.785 & 0.723 & 0.840 & 0.698 & 0.395 & 0.288 & 0.453 & 0.366 \\
& 90 & 0.760 & 0.700 & 0.815 & 0.675 & 0.360 & 0.252 & 0.418 & 0.330 \\
& 120 & 0.720 & 0.655 & 0.770 & 0.630 & 0.310 & 0.205 & 0.365 & 0.275 \\
\midrule
 & 30 & 0.777 & 0.745 & 0.859 & 0.730 & 0.420 & 0.322 & 0.468 & 0.394 \\
& 45 & 0.785 & 0.750 & 0.865 & 0.737 & 0.416 & 0.318 & 0.464 & 0.390 \\
LLaSa \cite{imran2024llasa} & 60 & 0.765 & 0.732 & 0.845 & 0.718 & 0.400 & 0.298 & 0.448 & 0.372 \\
& 90 & 0.740 & 0.708 & 0.820 & 0.690 & 0.365 & 0.264 & 0.415 & 0.339 \\
& 120 & 0.700 & 0.665 & 0.775 & 0.645 & 0.315 & 0.219 & 0.365 & 0.288 \\
\midrule
 & 30 & \textbf{0.814} & \textbf{0.855} & \textbf{0.893} & \textbf{0.843} & \textbf{0.478} & \textbf{0.410} & \textbf{0.565} & \textbf{0.522} \\
& 45 & \textbf{0.819} & \textbf{0.860} & \textbf{0.897} & \textbf{0.847} & \textbf{0.475} & \textbf{0.407} & \textbf{0.562} & \textbf{0.518} \\
\textbf{\sysName (ours)} & 60 & \textbf{0.808} & \textbf{0.848} & \textbf{0.887} & \textbf{0.838} & \textbf{0.457} & \textbf{0.392} & \textbf{0.544} & \textbf{0.505} \\
& 90 & \textbf{0.789} & \textbf{0.829} & \textbf{0.868} & \textbf{0.819} & \textbf{0.427} & \textbf{0.359} & \textbf{0.514} & \textbf{0.477} \\
& 120 & \textbf{0.739} & \textbf{0.784} & \textbf{0.819} & \textbf{0.768} & \textbf{0.377} & \textbf{0.309} & \textbf{0.464} & \textbf{0.427} \\
\bottomrule
\end{tabular}
\end{table*}


Given that the four metrics, BLEU@4, ROUGE-L, METEOR, and CIDEr, primarily focus on evaluating linguistic quality and surface similarity, while the RMC metric is specifically designed to measure the depth of underlying action semantics and functional information comprehension, we simplify the presentation for clarity. To facilitate observation and comparative analysis, this study uses S-Avg (Semantic Metric Average), calculated as the arithmetic mean of BLEU@4, ROUGE-L, METEOR, and CIDEr, as a composite indicator for the overall quality of text generation. We retain the RMC (Response Meaning Consistency) metric to specifically assess the model’s ability to capture action semantics.



As shown in Table~\ref{tab:long-sequence-ablation}, this experiment evaluates the change in model performance across varying input sequence lengths (30s, 45s, 60s, 90s, 120s) to assess robustness when dealing with continuous and redundant long-term temporal data. Overall, the performance trend exhibits a clear modality dependence and a general decline in long-term performance. Notably, on the IMU datasets (only IMU from XRF V2,
UWash), model performance did not decrease at the 45-second mark; rather, it showed a slight improvement. This suggests that for a high signal-to-noise ratio modality like IMU, a moderately longer sequence provides beneficial additional context that effectively reduces ambiguity in action boundaries. Therefore, investigating how to leverage continuously longer sequences to further enhance IMU model performance presents a potentially fruitful research direction. Conversely, performance on the Wi-Fi datasets (only Wi-Fi from XRF V2,
WiFiTAD) demonstrates a continuous decline starting immediately from the 30-second mark. For all modalities, performance uniformly begins to decay after 45 seconds, confirming that the aggregation of noise and redundant information in long sequences creates substantial difficulty for feature extraction and semantic comprehension. For instance, on the highly challenging XRF V2 (Wi-Fi) dataset, TimeLLM's RMC metric significantly dropped from 0.390 at 45 seconds to 0.288 at 120 seconds, underscoring the severity of the long-sequence challenge.


Despite the performance degradation observed across all models, our proposed \sysName  model maintains the highest performance ceiling and demonstrates the strongest robustness across all long-sequence settings. At their respective peak performance points (45 seconds for IMU, 30 seconds for Wi-Fi), \sysName achieved an absolute lead in all metrics.

\section{Conclusion}\label{sec:conclusion}

We presented \sysName, a framework capable of generating natural language descriptions of human activities directly from raw IMU or Wi-Fi CSI signals. To address the challenges of semantic extraction and long-range temporal modeling across heterogeneous modalities, we propose a modality-agnostic encoder that integrates patching mechanisms, placement embeddings, and Conv-FFN modules. Experimental results on the XRF V2, UWash, and WiFiTAD datasets fully validate the significant potential of signal-based captioning for enabling privacy-preserving, interpretable, and device-free activity monitoring in real-world settings.

Despite these advances, \sysName has only been evaluated on daily activity data from the aforementioned datasets and has not yet been tested in broader contexts, such as industrial environments (e.g., worker behavior analysis in smart factories), or clinical healthcare (e.g., postoperative rehabilitation motion description).  Future work can construct captioning datasets from industrial assembly lines, hospital wards, or athletic venues to systematically evaluate the model's generalization in more high-value scenarios.

\ifCLASSOPTIONcaptionsoff
  \newpage
\fi




\bibliographystyle{IEEEtran}
\bibliography{refrence}

@inproceedings{fan2020home,
  title={In-home daily-life captioning using radio signals},
  author={Fan, Lijie and Li, Tianhong and Yuan, Yuan and Katabi, Dina},
  booktitle={Computer Vision--ECCV 2020: 16th European Conference, Glasgow, UK, August 23--28, 2020, Proceedings, Part II 16},
  pages={105--123},
  year={2020},
  organization={Springer}
}

@inproceedings{anguita2013public,
  title={A Public Domain Dataset for Human Activity Recognition Using Smartphones},
  author={Anguita, Davide and Ghio, Alessandro and Oneto, Luca and Parra, Xavier and Reyes-Ortiz, Jorge L},
  booktitle={Proceedings of the 21st European Symposium on Artificial Neural Networks, Computational Intelligence and Machine Learning (ESANN)},
  pages={437--442},
  year={2013}
}

@inproceedings{gong2023mmg,
  title={MMG-Ego4D: Multi-Modal Generalization in Egocentric Action Recognition},
  author={Gong, Bing and Sikka, Karan and Huang, De-An and Grauman, Kristen},
  booktitle={Proceedings of the IEEE/CVF Conference on Computer Vision and Pattern Recognition (CVPR)},
  year={2023}
}

@inproceedings{li2022egopat3d,
  title={Egocentric Prediction of Action Target in 3D},
  author={Li, Yuxuan and Huang, De-An and Grauman, Kristen},
  booktitle={Proceedings of the European Conference on Computer Vision (ECCV)},
  year={2022}
}

@inproceedings{yi2022physical,
  title={Physical Inertial Poser (PIP): Physics-aware Real-time Human Motion Tracking from Sparse Inertial Sensors},
  author={Yi, Xinyu and Zhou, Yuxiao and Habermann, Marc and Shimada, Soshi and Golyanik, Vladislav and Theobalt, Christian and Xu, Feng},
  booktitle={Proceedings of the IEEE/CVF Conference on Computer Vision and Pattern Recognition (CVPR)},
  year={2022}
}

@inproceedings{von2017sparse,
  title={Sparse Inertial Poser: Automatic 3D Human Pose Estimation from Sparse IMUs},
  author={von Marcard, Timo and Henschel, Roberto and Black, Michael J and Rosenhahn, Bodo and Pons-Moll, Gerard},
  booktitle={Computer Graphics Forum (Proc. Eurographics)},
  volume={36},
  number={2},
  pages={349--360},
  year={2017}
}

@article{bock2024temporal,
  title={Temporal Action Localization for Inertial-based Human Activity Recognition},
  author={Bock, Marius and Moeller, Michael and Van Laerhoven, Kristof},
  journal={Proceedings of the ACM on Interactive, Mobile, Wearable and Ubiquitous Technologies},
  volume={8},
  number={4},
  pages={174:1--174:19},
  year={2024},
  doi={10.1145/3699770}
}

@article{lan2025xrf,
  author = {Lan, Bo and Li, Pei and Yin, Jiaxi and Song, Yunpeng and Wang, Ge and Ding, Han and Han, Jinsong and Wang, Fei},
  title = {XRF V2: A Dataset for Action Summarization with Wi-Fi Signals, and IMUs in Phones, Watches, Earbuds, and Glasses},
  journal = {Proceedings of the ACM on Interactive, Mobile, Wearable and Ubiquitous Technologies},
  volume={9},
  number={3},
  pages={1--41},
  year = {2025},
  publisher = {ACM}
}

@inproceedings{venugopalan2015sequence,
  title={Sequence to sequence-video to text},
  author={Venugopalan, Subhashini and Rohrbach, Marcus and Donahue, Jeffrey and Mooney, Raymond and Darrell, Trevor and Saenko, Kate},
  booktitle={Proceedings of the IEEE international conference on computer vision},
  pages={4534--4542},
  year={2015}
}

@inproceedings{yao2015describing,
  title={Describing videos by exploiting temporal structure},
  author={Yao, Li and Torabi, Atousa and Cho, Kyunghyun and Ballas, Nicolas and Pal, Christopher and Larochelle, Hugo and Courville, Aaron},
  booktitle={Proceedings of the IEEE international conference on computer vision},
  pages={4507--4515},
  year={2015}
}

@inproceedings{zhou2018end,
  title={End-to-end dense video captioning with masked transformer},
  author={Zhou, Luowei and Zhou, Yingbo and Corso, Jason J and Socher, Richard and Xiong, Caiming},
  booktitle={Proceedings of the IEEE conference on computer vision and pattern recognition},
  pages={8739--8748},
  year={2018}
}

@article{yang2021clip,
  title={Clip meets video captioners: Attribute-aware representation learning promotes accurate captioning},
  author={Yang, Bang and Zou, Yuexian},
  journal={CoRR},
  year={2021}
}

@inproceedings{tang2021clip4caption,
  title={Clip4caption: Clip for video caption},
  author={Tang, Mingkang and Wang, Zhanyu and Liu, Zhenhua and Rao, Fengyun and Li, Dian and Li, Xiu},
  booktitle={Proceedings of the 29th ACM International Conference on Multimedia},
  pages={4858--4862},
  year={2021}
}

@article{tang2021clip4caption++,
  title={Clip4caption++: Multi-clip for video caption},
  author={Tang, Mingkang and Wang, Zhanyu and Zeng, Zhaoyang and Rao, Fengyun and Li, Dian},
  journal={arXiv preprint arXiv:2110.05204},
  year={2021}
}

@article{shi2023learning,
  title={Learning video-text aligned representations for video captioning},
  author={Shi, Yaya and Xu, Haiyang and Yuan, Chunfeng and Li, Bing and Hu, Weiming and Zha, Zheng-Jun},
  journal={ACM Transactions on Multimedia Computing, Communications and Applications},
  volume={19},
  number={2},
  pages={1--21},
  year={2023},
  publisher={ACM New York, NY}
}

@article{nie2022time,
  title={A time series is worth 64 words: Long-term forecasting with transformers},
  author={Nie, Yuqi and Nguyen, Nam H and Sinthong, Phanwadee and Kalagnanam, Jayant},
  journal={arXiv preprint arXiv:2211.14730},
  year={2022}
}

@inproceedings{luo2024moderntcn,
  title={Moderntcn: A modern pure convolution structure for general time series analysis},
  author={Luo, Donghao and Wang, Xue},
  booktitle={The twelfth international conference on learning representations},
  pages={1--43},
  year={2024}
}

@inproceedings{zhang2021open,
  title={Open-book video captioning with retrieve-copy-generate network},
  author={Zhang, Ziqi and Qi, Zhongang and Yuan, Chunfeng and Shan, Ying and Li, Bing and Deng, Ying and Hu, Weiming},
  booktitle={Proceedings of the IEEE/CVF conference on computer vision and pattern recognition},
  pages={9837--9846},
  year={2021}
}

@article{vaswani2017attention,
  title={Attention is all you need},
  author={Vaswani, Ashish and Shazeer, Noam and Parmar, Niki and Uszkoreit, Jakob and Jones, Llion and Gomez, Aidan N and Kaiser, {\L}ukasz and Polosukhin, Illia},
  journal={Advances in neural information processing systems},
  volume={30},
  year={2017}
}

@inproceedings{islam2024video,
  title={Video recap: Recursive captioning of hour-long videos},
  author={Islam, Md Mohaiminul and Ho, Ngan and Yang, Xitong and Nagarajan, Tushar and Torresani, Lorenzo and Bertasius, Gedas},
  booktitle={Proceedings of the IEEE/CVF Conference on Computer Vision and Pattern Recognition},
  pages={18198--18208},
  year={2024}
}

@inproceedings{liu2025wifi,
  title={WiFi CSI Based Temporal Activity Detection via Dual Pyramid Network},
  author={Liu, Zhendong and Zhang, Le and Li, Bing and Zhou, Yingjie and Chen, Zhenghua and Zhu, Ce},
  booktitle={Proceedings of the AAAI Conference on Artificial Intelligence},
  volume={39},
  number={1},
  pages={550--558},
  year={2025}
}

@article{wang2025you,
  title={You Can Wash Hands Better: Accurate Daily Handwashing Assessment with a Smartwatch},
  author={Wang, Fei and Zhang, Tingting and Wu, Xilei and Wang, Pengcheng and Wang, Xin and Ding, Han and Shi, Jingang and Han, Jinsong and Huang, Dong},
  journal={IEEE Transactions on Mobile Computing},
  year={2025},
  publisher={IEEE}
}

@article{sun2024multimodal,
  title={Multimodal daily-life logging in free-living environment using non-visual egocentric sensors on a smartphone},
  author={Sun, Ke and Xia, Chunyu and Zhang, Xinyu and Chen, Hao and Zhang, Charlie Jianzhong},
  journal={Proceedings of the ACM on Interactive, Mobile, Wearable and Ubiquitous Technologies},
  volume={8},
  number={1},
  pages={1--32},
  year={2024},
  publisher={ACM New York, NY, USA}
}

@article{jin2023time,
  title={Time-llm: Time series forecasting by reprogramming large language models},
  author={Jin, Ming and Wang, Shiyu and Ma, Lintao and Chu, Zhixuan and Zhang, James Y and Shi, Xiaoming and Chen, Pin-Yu and Liang, Yuxuan and Li, Yuan-Fang and Pan, Shirui and others},
  journal={arXiv preprint arXiv:2310.01728},
  year={2023}
}

@misc{
li2024sensorllm,
title={Sensor{LLM}: Aligning Large Language Models with Motion Sensors for Human Activity Recognition},
author={Zechen Li and Shohreh Deldari and Linyao Chen and Hao Xue and Flora D. Salim},
year={2025},
url={https://openreview.net/forum?id=cDd7kg9mkP}
}

@article{wang2024xrf55,
  title={XRF55: A Radio Frequency Dataset for Human Indoor Action Analysis},
  author={Wang, Fei and Lv, Yizhe and Zhu, Mengdie and Ding, Han and Han, Jinsong},
  journal={Proceedings of the ACM on Interactive, Mobile, Wearable and Ubiquitous Technologies},
  volume={8},
  number={1},
  pages={1--34},
  year={2024},
  publisher={ACM New York, NY, USA}
}

@inproceedings{radford2021learning,
  title={Learning transferable visual models from natural language supervision},
  author={Radford, Alec and Kim, Jong Wook and Hallacy, Chris and Ramesh, Aditya and Goh, Gabriel and Agarwal, Sandhini and Sastry, Girish and Askell, Amanda and Mishkin, Pamela and Clark, Jack and others},
  booktitle={International conference on machine learning},
  pages={8748--8763},
  year={2021},
  organization={PMLR}
}

@article{li2025wilife,
  title={WiLife: Long-Term Daily Status Monitoring and Habit Mining of the Elderly Leveraging Ubiquitous Wi-Fi Signals},
  author={Li, Shengjie and Liu, Zhaopeng and Lv, Qin and Zou, Yanyan and Zhang, Yue and Zhang, Daqing},
  journal={ACM Transactions on Computing for Healthcare},
  volume={6},
  number={1},
  pages={1--29},
  year={2025},
  publisher={ACM New York, NY}
}

@article{imran2024llasa,
  title={Llasa: Large multimodal agent for human activity analysis through wearable sensors},
  author={Imran, Sheikh Asif and Khan, Mohammad Nur Hossain and Biswas, Subrata and Islam, Bashima},
  journal={CoRR},
  year={2024}
}

@article{abdar2024review,
  title={A review of deep learning for video captioning},
  author={Abdar, Moloud and Kollati, Meenakshi and Kuraparthi, Swaraja and Pourpanah, Farhad and McDuff, Daniel and Ghavamzadeh, Mohammad and Yan, Shuicheng and Mohamed, Abduallah and Khosravi, Abbas and Cambria, Erik and others},
  journal={IEEE Transactions on Pattern Analysis and Machine Intelligence},
  year={2024},
  publisher={IEEE}
}

@article{liu2025survey,
  title={Survey of Dense Video Captioning: Techniques, Resources, and Future Perspectives},
  author={Liu, Zhandong and Song, Ruixia},
  journal={Applied Sciences},
  volume={15},
  number={9},
  pages={4990},
  year={2025},
  publisher={MDPI}
}

@article{rafiq2023video,
  title={Video description: A comprehensive survey of deep learning approaches},
  author={Rafiq, Ghazala and Rafiq, Muhammad and Choi, Gyu Sang},
  journal={Artificial Intelligence Review},
  volume={56},
  number={11},
  pages={13293--13372},
  year={2023},
  publisher={Springer}
}

@inproceedings{pan2016hierarchical,
  title={Hierarchical recurrent neural encoder for video representation with application to captioning},
  author={Pan, Pingbo and Xu, Zhongwen and Yang, Yi and Wu, Fei and Zhuang, Yueting},
  booktitle={Proceedings of the IEEE conference on computer vision and pattern recognition},
  pages={1029--1038},
  year={2016}
}

@inproceedings{zhang2019object,
  title={Object-aware aggregation with bidirectional temporal graph for video captioning},
  author={Zhang, Junchao and Peng, Yuxin},
  booktitle={Proceedings of the IEEE/CVF conference on computer vision and pattern recognition},
  pages={8327--8336},
  year={2019}
}

@inproceedings{wang2019controllable,
  title={Controllable video captioning with pos sequence guidance based on gated fusion network},
  author={Wang, Bairui and Ma, Lin and Zhang, Wei and Jiang, Wenhao and Wang, Jingwen and Liu, Wei},
  booktitle={Proceedings of the IEEE/CVF international conference on computer vision},
  pages={2641--2650},
  year={2019}
}

@inproceedings{chen2019motion,
  title={Motion guided spatial attention for video captioning},
  author={Chen, Shaoxiang and Jiang, Yu-Gang},
  booktitle={Proceedings of the AAAI conference on artificial intelligence},
  volume={33},
  number={01},
  pages={8191--8198},
  year={2019}
}

@inproceedings{zhang2020object,
  title={Object relational graph with teacher-recommended learning for video captioning},
  author={Zhang, Ziqi and Shi, Yaya and Yuan, Chunfeng and Li, Bing and Wang, Peijin and Hu, Weiming and Zha, Zheng-Jun},
  booktitle={Proceedings of the IEEE/CVF conference on computer vision and pattern recognition},
  pages={13278--13288},
  year={2020}
}

@inproceedings{zheng2020syntax,
  title={Syntax-aware action targeting for video captioning},
  author={Zheng, Qi and Wang, Chaoyue and Tao, Dacheng},
  booktitle={Proceedings of the IEEE/CVF conference on computer vision and pattern recognition},
  pages={13096--13105},
  year={2020}
}

@inproceedings{pan2020spatio,
  title={Spatio-temporal graph for video captioning with knowledge distillation},
  author={Pan, Boxiao and Cai, Haoye and Huang, De-An and Lee, Kuan-Hui and Gaidon, Adrien and Adeli, Ehsan and Niebles, Juan Carlos},
  booktitle={Proceedings of the IEEE/CVF conference on computer vision and pattern recognition},
  pages={10870--10879},
  year={2020}
}

@inproceedings{lin2022swinbert,
  title={Swinbert: End-to-end transformers with sparse attention for video captioning},
  author={Lin, Kevin and Li, Linjie and Lin, Chung-Ching and Ahmed, Faisal and Gan, Zhe and Liu, Zicheng and Lu, Yumao and Wang, Lijuan},
  booktitle={Proceedings of the IEEE/CVF conference on computer vision and pattern recognition},
  pages={17949--17958},
  year={2022}
}

@article{tewel2022zero,
  title={Zero-shot video captioning with evolving pseudo-tokens},
  author={Tewel, Yoad and Shalev, Yoav and Nadler, Roy and Schwartz, Idan and Wolf, Lior},
  journal={arXiv preprint arXiv:2207.11100},
  year={2022}
}

@inproceedings{seo2022end,
  title={End-to-end generative pretraining for multimodal video captioning},
  author={Seo, Paul Hongsuck and Nagrani, Arsha and Arnab, Anurag and Schmid, Cordelia},
  booktitle={Proceedings of the IEEE/CVF conference on computer vision and pattern recognition},
  pages={17959--17968},
  year={2022}
}

@inproceedings{li2023lavender,
  title={Lavender: Unifying video-language understanding as masked language modeling},
  author={Li, Linjie and Gan, Zhe and Lin, Kevin and Lin, Chung-Ching and Liu, Zicheng and Liu, Ce and Wang, Lijuan},
  booktitle={Proceedings of the IEEE/CVF Conference on Computer Vision and Pattern Recognition},
  pages={23119--23129},
  year={2023}
}

@inproceedings{ji2024hargpt,
  title={Hargpt: Are llms zero-shot human activity recognizers?},
  author={Ji, Sijie and Zheng, Xinzhe and Wu, Chenshu},
  booktitle={2024 IEEE International Workshop on Foundation Models for Cyber-Physical Systems \& Internet of Things (FMSys)},
  pages={38--43},
  year={2024},
  organization={IEEE}
}

@inproceedings{hong2025llm4har,
  title={Llm4har: Generalizable on-device human activity recognition with pretrained llms},
  author={Hong, Zhiqing and Song, Yiwei and Li, Zelong and Yu, Anlan and Zhong, Shuxin and Ding, Yi and He, Tian and Zhang, Desheng},
  booktitle={Proceedings of the 31st ACM SIGKDD Conference on Knowledge Discovery and Data Mining V. 2},
  pages={4511--4521},
  year={2025}
}

@inproceedings{xu2024penetrative,
  title={Penetrative ai: Making llms comprehend the physical world},
  author={Xu, Huatao and Han, Liying and Yang, Qirui and Li, Mo and Srivastava, Mani},
  booktitle={Proceedings of the 25th International Workshop on Mobile Computing Systems and Applications},
  pages={1--7},
  year={2024}
}

@inproceedings{wang2025ego4o,
  title={Ego4o: Egocentric Human Motion Capture and Understanding from Multi-Modal Input},
  author={Wang, Jian and Dabral, Rishabh and Luvizon, Diogo and Cao, Zhe and Liu, Lingjie and Beeler, Thabo and Theobalt, Christian},
  booktitle={Proceedings of the Computer Vision and Pattern Recognition Conference},
  pages={22668--22679},
  year={2025}
}

@article{zhang2024unimts,
  title={UniMTS: Unified Pre-training for Motion Time Series},
  author={Zhang, Xiyuan and Teng, Diyan and Chowdhury, Ranak Roy and Li, Shuheng and Hong, Dezhi and Gupta, Rajesh and Shang, Jingbo},
  journal={Advances in Neural Information Processing Systems},
  volume={37},
  pages={107469--107493},
  year={2024}
}

@article{bock2024wear,
  title={Wear: An outdoor sports dataset for wearable and egocentric activity recognition},
  author={Bock, Marius and Kuehne, Hilde and Van Laerhoven, Kristof and Moeller, Michael},
  journal={Proceedings of the ACM on Interactive, Mobile, Wearable and Ubiquitous Technologies},
  volume={8},
  number={4},
  pages={1--21},
  year={2024},
  publisher={ACM New York, NY, USA}
}

@article{kim2024health,
  title={Health-llm: Large language models for health prediction via wearable sensor data},
  author={Kim, Yubin and Xu, Xuhai and McDuff, Daniel and Breazeal, Cynthia and Park, Hae Won},
  journal={arXiv preprint arXiv:2401.06866},
  year={2024}
}

@inproceedings{pu2013whole,
  title={Whole-home gesture recognition using wireless signals},
  author={Pu, Qifan and Gupta, Sidhant and Gollakota, Shyamnath and Patel, Shwetak},
  booktitle={Proceedings of the 19th annual international conference on Mobile computing \& networking},
  pages={27--38},
  year={2013}
}

@inproceedings{wang2015deepfi,
  title={DeepFi: Deep learning for indoor fingerprinting using channel state information},
  author={Wang, Xuyu and Gao, Lingjun and Mao, Shiwen and Pandey, Santosh},
  booktitle={2015 IEEE wireless communications and networking conference (WCNC)},
  pages={1666--1671},
  year={2015},
  organization={IEEE}
}

@inproceedings{wang2019person,
  title={Person-in-WiFi: Fine-grained person perception using WiFi},
  author={Wang, Fei and Zhou, Sanping and Panev, Stanislav and Han, Jinsong and Huang, Dong},
  booktitle={Proceedings of the IEEE/CVF international conference on computer vision},
  pages={5452--5461},
  year={2019}
}

@inproceedings{kotaru2015spotfi,
  title={Spotfi: Decimeter level localization using wifi},
  author={Kotaru, Manikanta and Joshi, Kiran and Bharadia, Dinesh and Katti, Sachin},
  booktitle={Proceedings of the 2015 ACM conference on special interest group on data communication},
  pages={269--282},
  year={2015}
}

@article{halperin2011tool,
  title={Tool release: Gathering 802.11 n traces with channel state information},
  author={Halperin, Daniel and Hu, Wenjun and Sheth, Anmol and Wetherall, David},
  journal={ACM SIGCOMM computer communication review},
  volume={41},
  number={1},
  pages={53--53},
  year={2011},
  publisher={ACM New York, NY, USA}
}

@article{wang2024multi,
  title={Multi-Subject 3D Human Mesh Construction Using Commodity WiFi},
  author={Wang, Yichao and Ren, Yili and Yang, Jie},
  journal={Proceedings of the ACM on Interactive, Mobile, Wearable and Ubiquitous Technologies},
  volume={8},
  number={1},
  pages={1--25},
  year={2024},
  publisher={ACM New York, NY, USA}
}

@inproceedings{zheng2019zero,
  title={Zero-effort cross-domain gesture recognition with Wi-Fi},
  author={Zheng, Yue and Zhang, Yi and Qian, Kun and Zhang, Guidong and Liu, Yunhao and Wu, Chenshu and Yang, Zheng},
  booktitle={Proceedings of the 17th annual international conference on mobile systems, applications, and services},
  pages={313--325},
  year={2019}
}

@article{chen2024sensor2text,
  title={Sensor2text: Enabling natural language interactions for daily activity tracking using wearable sensors},
  author={Chen, Wenqiang and Cheng, Jiaxuan and Wang, Leyao and Zhao, Wei and Matusik, Wojciech},
  journal={Proceedings of the ACM on Interactive, Mobile, Wearable and Ubiquitous Technologies},
  volume={8},
  number={4},
  pages={1--26},
  year={2024},
  publisher={ACM New York, NY, USA}
}

@article{howard2017mobilenets,
  title={Mobilenets: Efficient convolutional neural networks for mobile vision applications},
  author={Howard, Andrew G and Zhu, Menglong and Chen, Bo and Kalenichenko, Dmitry and Wang, Weijun and Weyand, Tobias and Andreetto, Marco and Adam, Hartwig},
  journal={arXiv preprint arXiv:1704.04861},
  year={2017}
}

@ARTICLE{10556745,
  author={He, Yinghui and Liu, Jianwei and Li, Mo and Yu, Guanding and Han, Jinsong},
  journal={IEEE J. Sel. Areas Commun.}, 
  title={{Forward-compatible integrated sensing and communication for WiFi}}, 
  year={2024},
  volume={42},
  number={9},
  pages={2440--2456},
  month={Sep.}
}
%

%

\begin{IEEEbiography}[{\includegraphics[width=0.8in,clip,keepaspectratio]{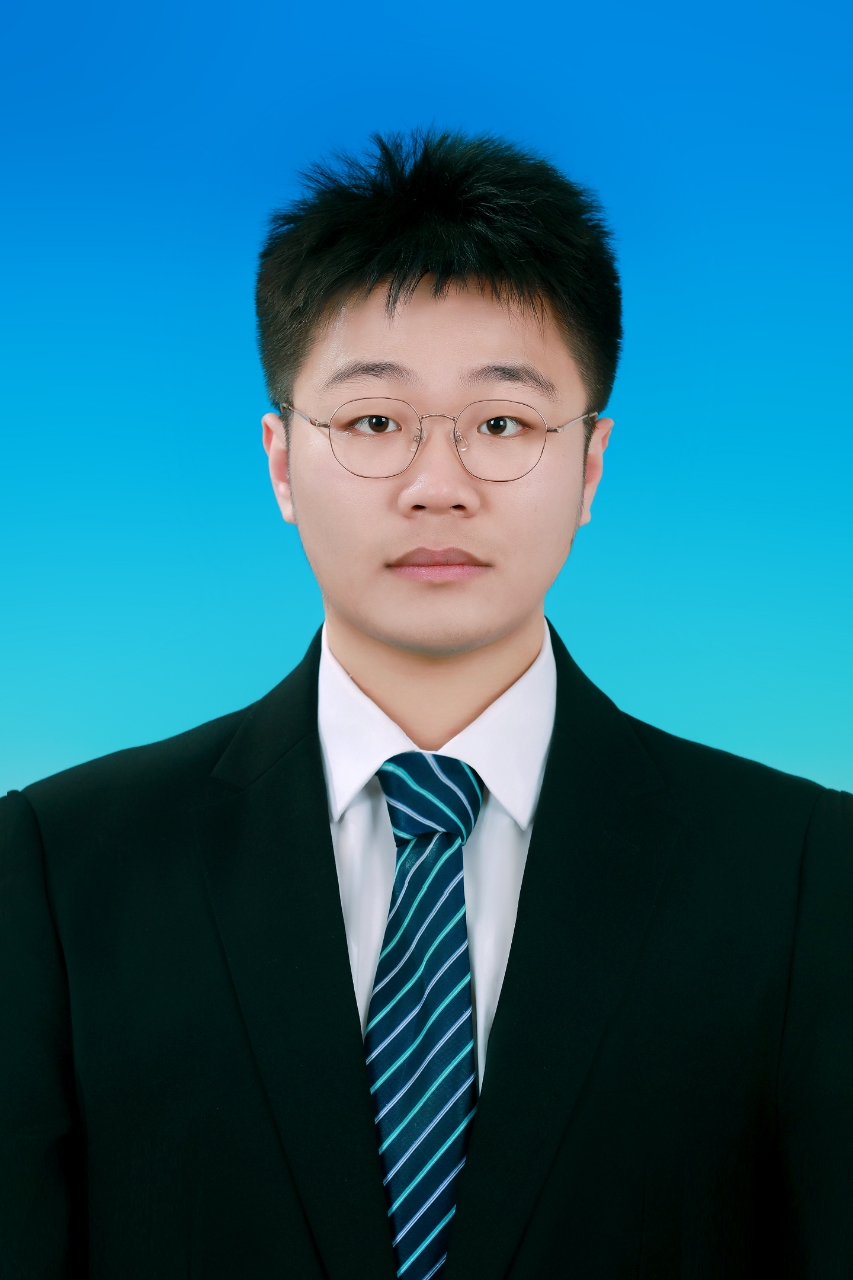}}]
{Fei Deng} received Bachelor's degree in Software Engineering from Hunan University, China, in 2024. He is currently a master student at Xi'an Jiaotong University, China. His research interests include mobile computing and time series processing.
\end{IEEEbiography}

\vspace{-32pt}
\begin{IEEEbiography}[{\includegraphics[width=0.8in,clip]{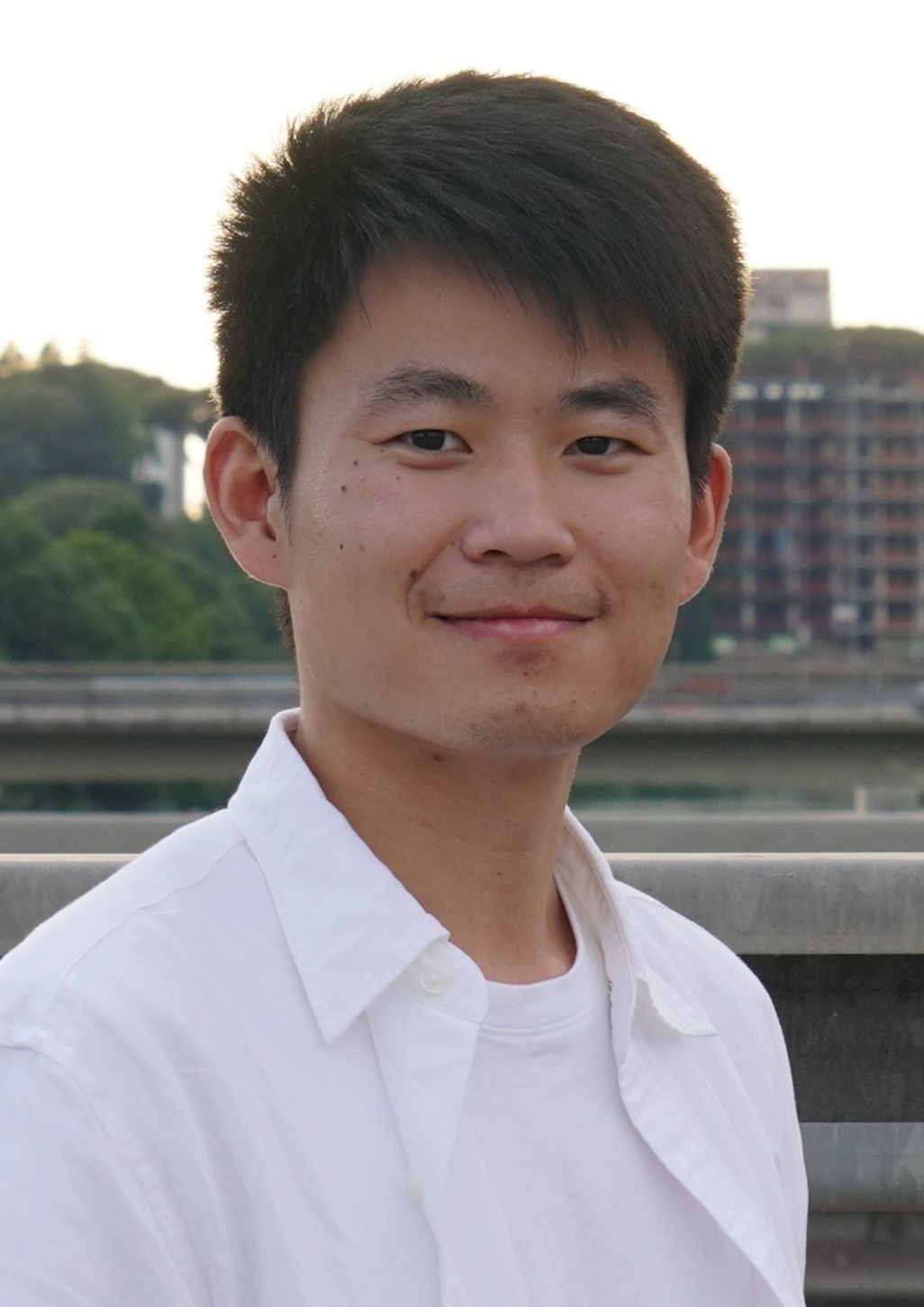}}]{Yinghui He} (Member, IEEE) received the B.E. degree in information engineering and Ph.D. degree in information and communication engineering from Zhejiang University, Hangzhou, China, in 2018 and 2023, respectively. He is currently a Research Fellow with the College of Computing and Data Science, Nanyang Technological University, Singapore. 
His research interests mainly include integrated sensing and communications (ISAC), mobile computing, and device-to-device communications.
\end{IEEEbiography}

\vspace{-32pt}
\begin{IEEEbiography}[{\includegraphics[width=0.8in,clip,keepaspectratio]{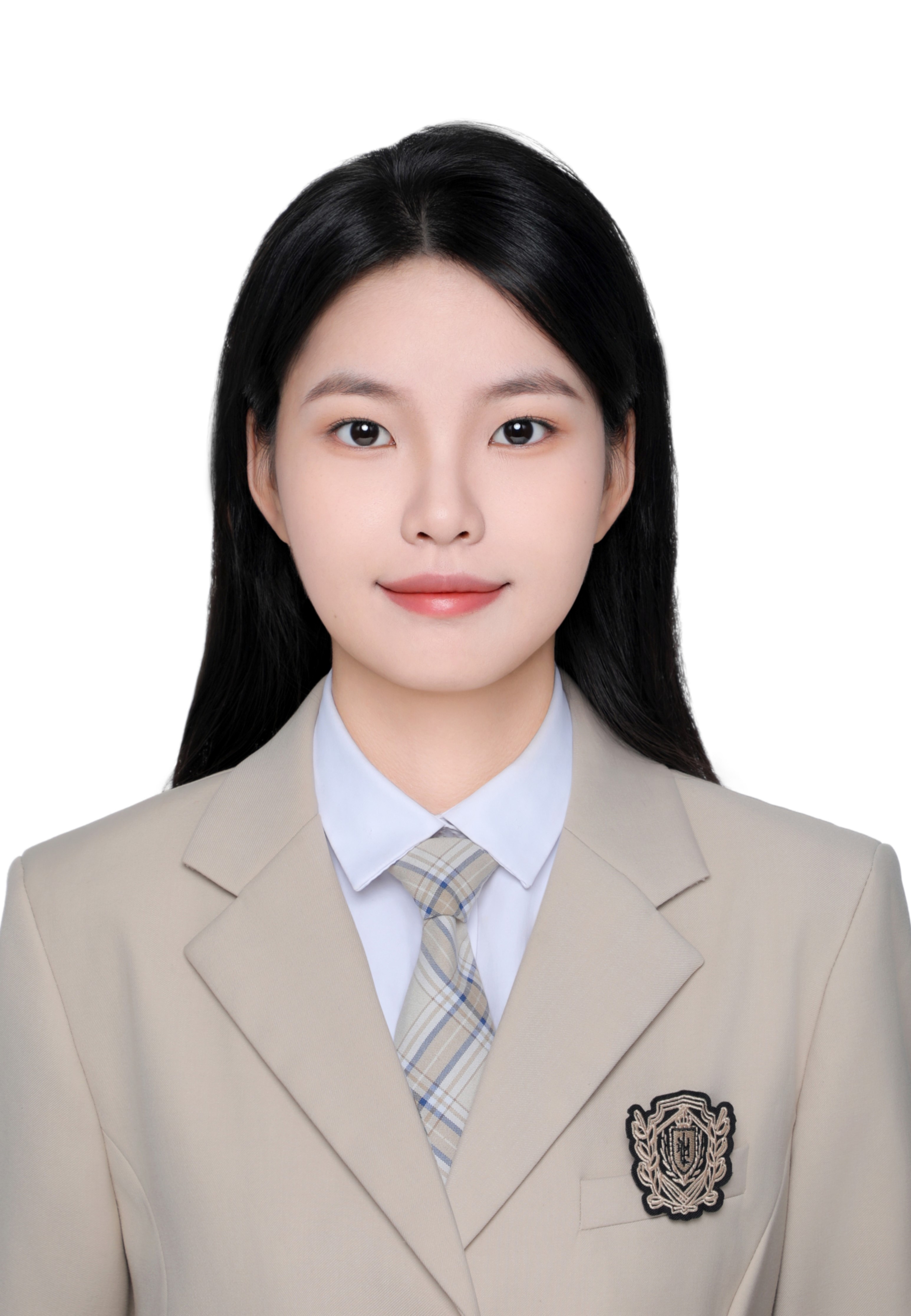}}]
{Chuntong Chu} received Bachelor's degree in Information Management and Information Systems from Northeastern University, China, in 2025. She is currently pursuing a Master's degree in Software Engineering at the School of Software, Xi'an Jiaotong University, Xi'an, China. Her research interests focus on AIoT,and Human-centered AI,Ubiquitous sensing
\end{IEEEbiography}

\vspace{-32pt}
\begin{IEEEbiography}[{\includegraphics[width=0.8in,clip,keepaspectratio]{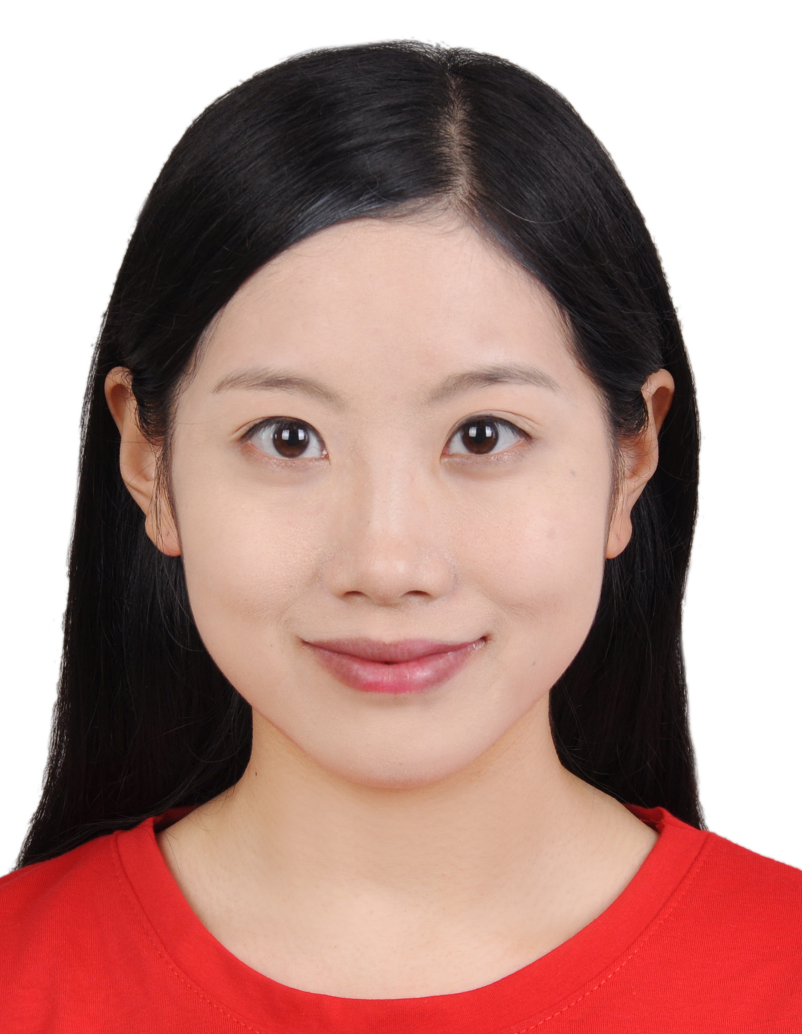}}]
{Ge Wang} (Member, IEEE) is now an Associate Professor at Xi'an Jiaotong University. She received her Ph.D. at Xi'an Jiaotong University in 2019. Her research interests include wireless sensor network and mobile computing.
\end{IEEEbiography}

\vspace{-32pt}
\begin{IEEEbiography}[{\includegraphics[width=0.8in,clip,keepaspectratio]{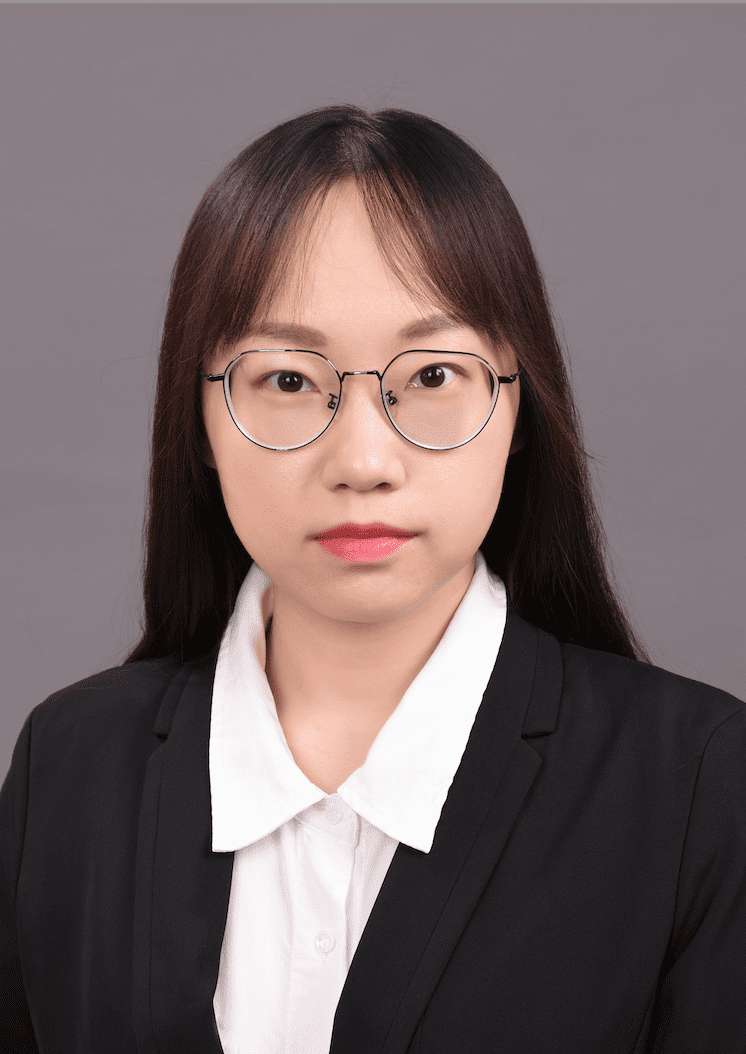}}]
{Han Ding} (Senior Member, IEEE) received the Ph.D. degree in computer science and technology from Xi’an Jiaotong University, Xi’an, China, in 2017.
She is now a professor with Xi’an Jiaotong University. His research interests focus on AIoT, smart sensing, and RFID systems.
\end{IEEEbiography}
\vspace{-32pt}

\begin{IEEEbiography}[{\includegraphics[width=0.8in,clip,keepaspectratio]{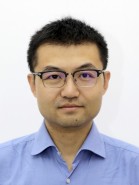}}]
{Jinsong Han} (Senior Member, IEEE) received his Ph.D. degree from Hong Kong University of Science and Technology in 2007. He is now a professor with Zhejiang University, Hangzhou, China. He is a senior member of the ACM and IEEE. His research interests focus on IoT security, smart sensing, wireless and mobile computing.
\end{IEEEbiography}

\vspace{-32pt}
\begin{IEEEbiography}[{\includegraphics[width=0.8in,clip,keepaspectratio]{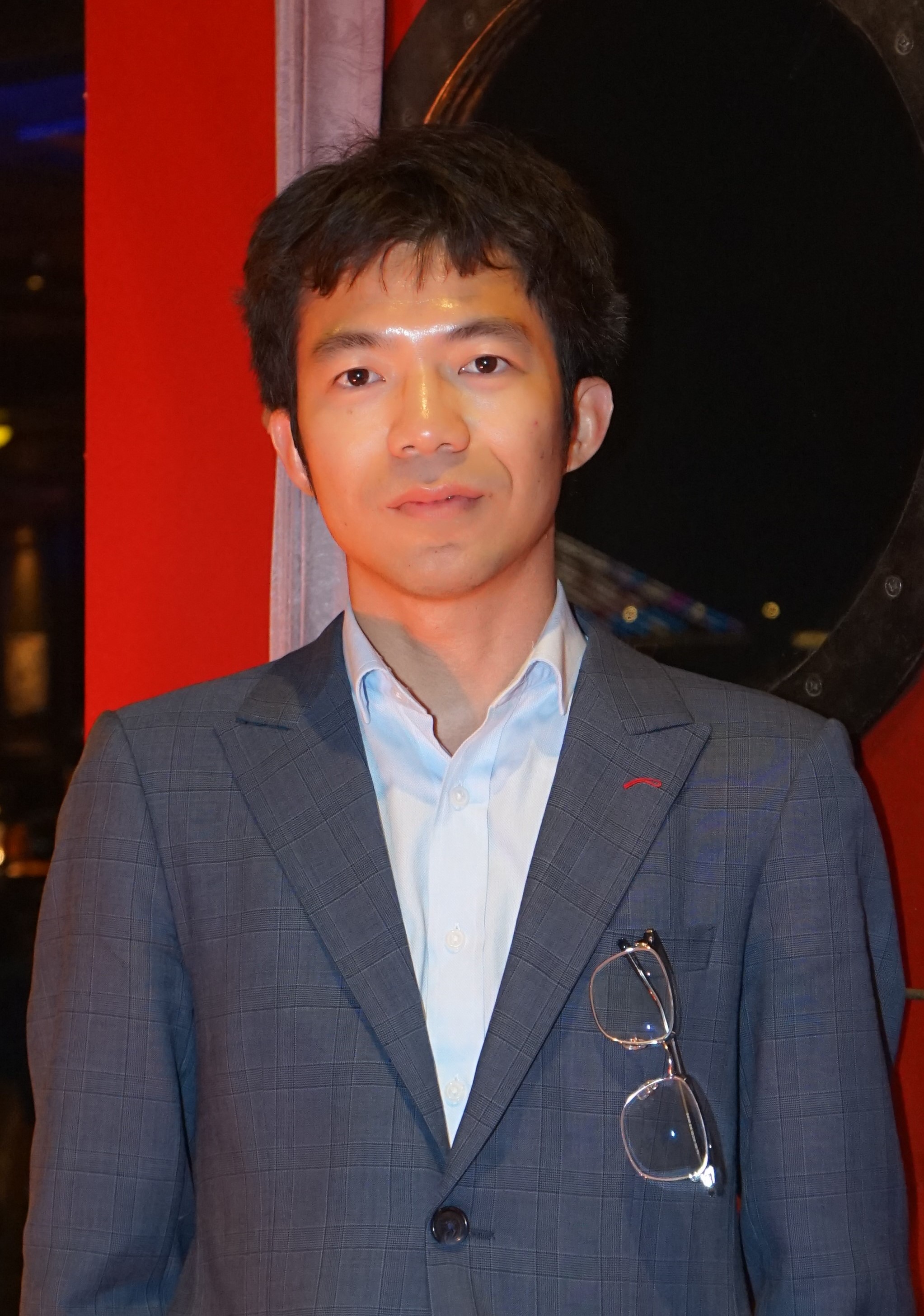}}]
{Fei Wang} (Member, IEEE) received the B.E. and Ph.D. degrees in Computer Science and Technology from Xi’an Jiaotong University, Xi'an, China, in 2013 and 2020, respectively. He was a visiting Ph.D. student with the School of Computer Science, Carnegie Mellon University, Pittsburgh, USA, from 2017 to 2019. He is currently an associate professor with Xi’an Jiaotong University. His research interests include human sensing, mobile computing, and deep learning.
\end{IEEEbiography}




\end{document}